%% file: shellfishnet_arxiv.tex
\documentclass[10pt]{article}
\usepackage{arxiv}
\usepackage{arxiv-consistent}
\usepackage{longtable}

\input{math_commands.tex}

\newcommand{\hcenteredincludegraphics}[4][]{%
  \makebox[\linewidth][c]{\hspace*{#2}\includegraphics[#1]{#4}\hspace*{#3}}}

\title{ShellfishNet: A Domain-Specific Benchmark for Visual Recognition of Marine Molluscs}
\renewcommand{\shorttitle}{ShellfishNet}

\makeatletter
\gdef\@author{%
  \parbox[t]{0.45\textwidth}{\centering
      {\Authfont Ziheng Zhou}\\
      {\Affilfont IT College,}\\
      {\Affilfont Shanghai Ocean University}\\
      {\Affilfont\ttfamily zihengzhouac@outlook.com}
  }\hspace{0.04\textwidth}%
  \parbox[t]{0.45\textwidth}{\centering
      {\Authfont Yang Wang}\\
      {\Affilfont Fudan University}\\
      {\Affilfont\ttfamily fdwy@fudan.edu.cn}
  }\\[2.2em]
  \parbox[t]{0.30\textwidth}{\centering
      {\Authfont Nan Wang}\\
      {\Affilfont DP Technology}\\
      {\Affilfont\ttfamily wangnan411570@gmail.com}
  }\hspace{0.02\textwidth}%
  \parbox[t]{0.30\textwidth}{\centering
      {\Authfont Chengliang Wu}\\
      {\Affilfont IT College,}\\
      {\Affilfont Shanghai Ocean University}\\
      {\Affilfont\ttfamily clwu@shou.edu.cn}
  }\hspace{0.02\textwidth}%
  \parbox[t]{0.30\textwidth}{\centering
      {\Authfont Jun Yan\textsuperscript{*}}\\
      {\Affilfont IT College,}\\
      {\Affilfont Shanghai Ocean University}\\
      {\Affilfont\ttfamily yanjun@ieee.org}
  }%
}
\makeatother
\date{}

\hypersetup{
  pdftitle={ShellfishNet: A Domain-Specific Benchmark for Visual Recognition of Marine Molluscs},
  pdfauthor={Ziheng Zhou, Yang Wang, Nan Wang, Chengliang Wu, Jun Yan},
  pdfkeywords={shellfish recognition, marine vision, fine-grained classification, image retrieval}
}

\begin{document}
\maketitle

\begin{abstract}
Reliable visual recognition of shellfish is important for ecological monitoring and aquaculture, but real-world imagery exhibits substantial variation in acquisition conditions, appearance, and posture. Existing marine vision resources cover diverse taxa and imaging settings, but fine-grained mollusc recognition remains underexplored in a group-disjoint benchmark that jointly evaluates class prediction, representation-space retrieval, and condition sensitivity. To address this gap, we introduce ShellfishNet, a provenance-aware, group-disjoint benchmark of 40,130 images from 100 marine mollusc classes, organized into 30,879 acquisition or observation groups. Rather than treating recognition as a single clean-accuracy leaderboard, ShellfishNet evaluates model configurations across classification, image-to-image retrieval, taxonomic error structure, and controlled condition sensitivity. We benchmark representative convolutional, transformer-based, hybrid and efficient, self-supervised, and vision-language models on classification and retrieval. Separately, we evaluate open image-captioning and vision-language models on the captioning subset and assess selected fixed checkpoints under controlled generic and marine-inspired synthetic transformations. Across the benchmark, image-level and class-aware classification largely agree, whereas retrieval provides complementary model rankings and taxonomy projection reveals that a non-negligible subset of exact-class errors remains correct at higher taxonomic ranks. Selected fixed checkpoints with similar clean accuracy can also exhibit distinct sensitivity profiles under controlled transformations. These results show that model comparison on shellfish imagery should account for both task and image condition, rather than rely on a single clean classification score. The dataset and evaluation protocols will be publicly released.
\end{abstract}


\section{Introduction}

The management of nearshore waters and benthic ecosystems is important for maintaining biodiversity and supporting coastal economies. Detailed knowledge of shellfish can clarify their ecological and economic contributions and inform conservation efforts. Ecologically, a global synthesis found that structural restoration of oyster habitats was followed by rapid increases in the biodiversity and abundance of reef-associated fauna during the first two years, although recovery remained 35\% below predisturbance levels~\citep{hemraj2022oyster}. Economically, a global review estimated substantial market and non-market value associated with bivalve aquaculture, including food production and non-food ecosystem services~\citep{olivier2020bivalve}.

\input{fig/figure1_layout}

Maintaining ecological resilience requires early, proactive monitoring to enable timely interventions and adaptive strategies that sustain ecosystem functions. However, traditional benthic monitoring relies heavily on specialized taxonomic expertise for morphological identification. This labor-intensive and time-consuming process creates a bottleneck, inherently limiting its scalability for broad-scale ecological assessments~\citep{lejzerowicz2015benthic}. In recent years, rapid advances in computer vision technologies have created substantial opportunities for automated surveys in ecology~\citep{weinstein2018computer}. Specifically, foundation models have demonstrated broad transfer capabilities across visual recognition tasks~\citep{radford2021learning}. Unlike many common visual recognition targets, marine shellfish can exhibit high intra-class variation and subtle inter-class differences~\citep{zhang2019shell,valverde2025backhome}. Existing studies suggest that, when confronted with specialized biological taxa exhibiting high morphological similarity, general-purpose foundation models can show domain-specific limitations in fine-grained recognition~\citep{khan2025fishnetpp,felemban2025reefnet,yu2026fgbmk}. Such issues can restrict the utility of artificial intelligence technologies for marine ecological monitoring. Furthermore, complex and variable field conditions, including turbid water, dynamic lighting, challenging viewing angles, and complex backgrounds, can degrade image quality or obscure discriminative features, thereby increasing recognition difficulty~\citep{akkaynak2018revised,khan2023fishnet}.

High-quality, domain-specific benchmarks remain scarce. Existing shell-recognition resources include datasets acquired under standardized protocols, such as paired frontal and lateral specimen views, which capture less of the variation in natural lighting, habitat backgrounds, viewing angles, and postures found in field imagery~\citep{zhang2019shell}. Broader aquatic resources provide greater scale and visual diversity. FishNet contains 94,532 images spanning 17,357 aquatic species, incorporates varied image sizes, resolutions, and illumination conditions, and supports prediction of functional ecological traits such as habitat and nutritional value~\citep{khan2023fishnet}. Its emphasis, however, remains on fish rather than benthic shellfish. The resulting gap in taxonomic scope and imaging conditions motivates a shellfish-specific benchmark for fine-grained recognition across varied habitats and acquisition settings~\citep{zhang2019shell,valverde2025backhome,khan2023fishnet}. Together, these resources leave limited support for a fixed-label, group-disjoint mollusc benchmark unifying classification, retrieval, and controlled condition analysis.

To systematically evaluate the fine-grained recognition and description capabilities of existing vision and multimodal models across varied shellfish imaging conditions, we introduce a domain-specific benchmark. Fig.~\ref{Fig:figure1} presents an overview of ShellfishNet, from dataset construction to visual benchmark evaluation and image captioning. The benchmark targets visual variation relevant to shellfish identification across varied acquisition conditions and subject presentations. Comprising 40,130 images across 100 classes, ShellfishNet uses group-disjoint training, validation, and test partitions across 30,879 acquisition or observation groups. Based on the proposed dataset, we benchmark representative models for image classification and image-to-image retrieval and conduct a separate descriptive breadth scan of 80 backbones. The benchmark spans Convolutional Neural Networks~\citep{he2016deep}, Vision Transformers (ViTs)~\citep{dosovitskiy2021image}, hybrid and efficient architectures~\citep{tu2022maxvit}, self-supervised learning methods~\citep{oquab2024dinov2}, and vision-language encoders~\citep{cherti2023openclip}. Furthermore, to extend the benchmark beyond categorical classification, we provide an additional subset with descriptive text captions. Whereas discrete taxonomic labels specify class identity, natural-language descriptions can additionally express visible attributes and scene context. The subset supports comparison of 10 open image-captioning and vision-language models. Results show different classification and retrieval rankings and distinct responses to controlled transformations among fixed checkpoints with similar clean-image accuracy.

The main contributions of this work are as follows:
\begin{itemize}
    \item \textbf{A provenance-aware, group-disjoint benchmark for fine-grained mollusc recognition.} We construct ShellfishNet, an image dataset comprising 40,130 images across 100 classes, with group-disjoint training, validation, and test partitions across 30,879 acquisition or observation groups. The benchmark targets visual variation relevant to shellfish identification across controlled and intertidal scenes, with variation in image source and subject presentation.
\item \textbf{Detailed model evaluation.} We benchmark a diverse set of representative models on image classification and image-to-image retrieval using fixed data partitions, common evaluation metrics, and documented model-family training profiles. Furthermore, we assess the sensitivity of selected fixed checkpoints to controlled generic and marine-inspired synthetic transformations.
    \item \textbf{A human-anchored multimodal extension.} Beyond visual recognition, we use a subset with descriptive text annotations to explore the image-captioning capabilities of 10 open image-captioning and vision-language models. This exploration provides a common basis for comparing multimodal image description on shellfish imagery.
\end{itemize}

\section{Related Work}
\par Natural-world benchmarks such as CUB-200-2011~\citep{wah2011caltech}, iNaturalist~\citep{van2018inaturalist}, and Stanford Cars~\citep{krause20133d} established widely used settings for fine-grained visual recognition. Marine datasets extend these ideas to aquatic taxonomy and underwater imagery. FishNet contains 94,532 images spanning 17,357 aquatic species and supports recognition, detection, and functional-trait prediction~\citep{khan2023fishnet}. FathomNet aggregates expert-curated labels for marine animals and other underwater concepts~\citep{katija2022fathomnet}. BenthicNet compiles more than 11.4 million seafloor images and curates a representative subset of 1.3 million images for model development~\citep{lowe2025benthicnet}. These resources provide broad taxonomic or environmental coverage, but their tasks and annotations differ from those of a fixed mollusc recognition benchmark.

Recent marine benchmarks examine fine-grained categories, source variation, and multimodal supervision. ReefNet aggregates approximately 925,000 genus-level hard-coral annotations from 76 CoralNet sources and an additional Red Sea site. Its benchmark covers zero-shot recognition, cross-domain few-shot adaptation, within-source evaluation, and cross-source transfer~\citep{felemban2025reefnet}. FishNet++ evaluates multimodal large language models on fine-grained fish recognition. The best tested open-source model achieves less than 10\% accuracy, and the resource adds descriptions, keypoints, and bounding boxes~\citep{khan2025fishnetpp}. ORCA provides 14,647 images from 478 marine species, 42,217 bounding boxes, and 22,321 expert-verified instance captions. It evaluates object detection, instance captioning, and visual grounding~\citep{wong2026orca}.

Fine-grained model diagnosis also extends beyond marine datasets. FG-BMK evaluates large vision--language models at dialogue and feature levels across common and specialized domains. It identifies persistent fine-grained recognition failures and reports DINOv2 among the stronger encoders in feature-level retrieval and classification~\citep{yu2026fgbmk}. Together, FishNet++ and FG-BMK support direct comparison of language-aligned models and vision-only encoders in fine-grained evaluation.

Work focused on shells and molluscs remains more specialized. \citet{zhang2019shell} compiled 59,244 paired frontal and lateral images from 29,622 specimens representing 7,894 shell species. The standardized two-view protocol provides controlled specimen imagery, while later recognition methods address high visual similarity and transfer learning in shellfish and fish classification~\citep{zhang2023high,ergun2025swinfishnet}. BackHome19K contains 19,051 natural-light images across 516 mollusc species and targets coast-level geographic provenance~\citep{valverde2025backhome}. ShellfishNet complements these resources with group-disjoint splits for fine-grained mollusc classification and retrieval, a common protocol across model families, controlled transformation tests, and a 512-image caption subset. ShellfishNet emphasizes image-level taxonomic comparison. BackHome19K targets provenance, ReefNet studies coral genera and cross-source transfer, and ORCA evaluates instance-level detection and description. BenthicNet provides seafloor coverage at a much larger scale.

\input{fig/figure2_layout}

\section{Benchmark Pipeline Design}
\label{sec:benchmark_pipeline}

ShellfishNet organizes four complementary evaluations around group-disjoint data partitions. Closed-set classification measures category prediction, while image-to-image retrieval tests whether the same classifiers organize same-class images in their learned representation spaces. Captioning extends the benchmark from class identity to visible attributes and scene context. Controlled transformations compare fixed classifiers under prescribed changes in noise, blur, brightness, and contrast, as well as marine-inspired Jerlov II and III attenuation and marine snow.

\subsection{Dataset Construction and Split Design}

ShellfishNet draws on publicly accessible biodiversity and natural-history repositories. Sources include iNaturalist~\citep{van2018inaturalist}, GBIF~\citep{lane2007gbif}, Wikimedia Commons~\citep{vaidya2015dbpediacommons}, iDigBio~\citep{singer2018idigbio}, and the Encyclopedia of Life~\citep{parr2014eol}, among others. For each image, the release records the source platform, record and media references, URLs, attribution, and license. Images are distributed under CC0, CC BY, or CC BY-SA terms. The corresponding attribution and license links are retained as part of each image's provenance record. Fig.~\ref{Fig:figure2} shows representative scenes, partition sizes, class-size variation, and acquisition or observation group sizes.

Before inclusion, each record must have a traceable source URL and an eligible license. Images must have a shorter side of at least 512 pixels. We retain at most two images from one acquisition or observation group and at most 25 images per photographer within each class. An acquisition or observation group links images from the same capture event or source record. Group identifiers use an acquisition session, observation, or specimen identifier when available. Otherwise, they use the source platform and source record. Class labels are checked by a reviewer familiar with shellfish taxonomy, who verifies that the principal subject in each image matches its assigned class. Images that cannot be identified reliably, lack a clear principal subject, or contain conflicting subjects are marked as ambiguous. Scientific names follow the accepted names in WoRMS~\citep{worms2026}; outdated names and synonyms are mapped to their accepted names. Ambiguous cases and nomenclatural inconsistencies are rechecked against the original source record, image content, and taxonomic database. Samples that cannot be resolved are excluded. Each released image is assigned to the resulting fixed 100-class label set with a class identifier and scientific name. The benchmark contains 40,130 images across these classes. Each group is assigned wholly to one partition. The resulting partitions contain 32,143 training images, 4,029 validation images, and 3,958 test images across 30,879 unique groups. No group is shared between partitions, and the test partition remains reserved for evaluation. Class sizes range from 132 to 500 images. We report class-aware metrics alongside overall accuracy.

The classification and retrieval evaluations use these fixed partitions, while the captioning extension draws its fixed evaluation subset exclusively from the test partition. The collection retains variation in image source and subject presentation, while the released scene field supports controlled and intertidal stratification.

\subsection{Classification and Retrieval Benchmark}

The primary visual benchmark covers closed-set classification over 100 shellfish classes and class-based image-to-image retrieval. We compare representative visual models from convolutional, transformer-based, hybrid and efficient, self-supervised, and vision-language model families. Retrieval uses the 3,958 test images as queries and the 32,143 training images as a gallery. Images from the same class are treated as relevant.

A descriptive breadth scan evaluates 80 backbones using one fixed random seed and 30 training epochs. Appendix Secs.~\ref{app:breadth-scan} and~\ref{app:breadth-class-analysis} report the complete model-level and efficiency results and the per-class analysis, respectively.

\subsection{Captioning Extension}

The captioning extension adds descriptive output to categorical recognition. It uses a fixed subset of 512 test images that covers all 100 classes. Chat-based vision-language models receive a common English instruction, Florence-2 uses its native caption task prompt, and classic captioning models use their native image-to-text interfaces with deterministic generation. We compare 10 open image-captioning and vision-language models against a separate set of reference captions.

A human domain expert leads the reference-caption annotation and determines the final wording. GPT-5.6 Sol~\citep{openai2026gpt56} assists with initial drafts, after which the expert inspects each image and revises every caption to describe visible shellfish attributes and scene context within the concise one-sentence format.

\section{Experiments}
\label{sec:experiments}

We evaluate ShellfishNet through closed-set classification, image-to-image retrieval, captioning, and controlled transformations. We use fixed data partitions, specified input, and evaluation protocols.

\input{table/classification_main}

\input{fig/figure3_layout}

\subsection{Experimental Setup}
\label{sec:experimental_setup}

All classification models use the same 100-class training, validation, and test partitions. During training, random resized cropping produces $224\times224$-pixel images. The crop scale ranges from $0.80$ to $1.00$, with aspect ratios between $3/4$ and $4/3$. We apply horizontal flipping with probability $0.5$. Brightness, contrast, and saturation factors are sampled from $[0.90,1.10]$. Validation and test images are resized so that their shorter side is 256 pixels and then center-cropped to $224\times224$, without test-time augmentation. All classification models except OpenCLIP ViT-B/16~\citep{cherti2023openclip} and SigLIP2 Base patch16-224~\citep{tschannen2025siglip2} use ImageNet normalization. These two encoders use their released normalization statistics. Each model is trained for 100 epochs using cross-entropy loss without label smoothing. We use a cosine-annealing schedule updated after each epoch, with a minimum learning rate of zero. Validation Macro-F1 determines checkpoint selection. Training uses FP16 automatic mixed precision, while validation and test inference use FP32.

Optimizer settings follow two fixed architecture-specific profiles. Sixteen models use AdamW with learning rate $3\times10^{-5}$ and weight decay $0.05$. The remaining seven models use SGD with learning rate $0.01$, momentum $0.9$, and weight decay $0.0001$. We selected these profiles using validation performance before test evaluation. Every run has an effective batch size of 48, with gradient accumulation used where required by memory. Each model is trained with seeds 41, 42, and 43, yielding 69 formal runs. For each model, we aggregate held-out classification and retrieval results using the arithmetic mean and sample standard deviation over its runs. Table~\ref{tab:classification-main} reports the means for all seven measures, while Appendix Sec.~\ref{app:heldout-three-seed} reports sample standard deviations for all seven metrics and seed-specific Top-1 and class-macro mAP@R results.

For classification, Top-1 and Top-5 accuracy measure whether the true class is the highest-scoring prediction or appears among the five highest-scoring predictions, respectively. Macro-F1 and Balanced Accuracy average performance across classes and can reveal effects hidden by the overall image count. We report all four measures on the held-out test partition.

For image-to-image retrieval, the 3,958 test images serve as queries against a gallery of 32,143 training images. Images from the same class are relevant. For each classifier, retrieval uses the tensor supplied directly to its trained 100-class classification head. Query and gallery vectors are L2-normalized and ranked by cosine similarity, without additional retrieval training. We report Recall@1, Recall@5, and class-macro mAP@R~\citep{musgrave2020metric}. Here, $R$ is the number of same-class gallery images for a query. Appendix Sec.~\ref{app:retrieval-measures} reports other retrieval measures.

For captioning, the fixed subset contains 512 test images and covers all 100 classes. Each image is provided to all ten open models in the same standardized $1024\times1024$ RGB format. Chat-based vision-language models use a common scientific image-captioning instruction. Florence-2 uses its native caption task prompt, while the classic captioning models use their native image-to-text interfaces. The common instruction requests one English sentence of at most 30 words describing the visible subject, salient attributes, and marine scene context. It permits a common name, genus, or species only when supported by visible evidence; otherwise, it requests observable characteristics without speculation. Decoding uses batch size one and at most 48 new tokens. Prompted models use temperature zero where supported, while the classic models use deterministic generation through their native interfaces. The reference captions are annotated and finalized by a human domain expert, with GPT-5.6 Sol assisting with initial drafts. We compare generated descriptions with these references using CIDEr~\citep{vedantam2015cider}, SPICE~\citep{anderson2016spice}, CLIPScore~\citep{hessel2021clipscore}, and BERTScore~\citep{zhang2020bertscore}. We also report BLEU-4~\citep{papineni2002bleu}, METEOR~\citep{banerjee2005meteor}, ROUGE-L~\citep{lin2004rouge}, scientific-name token mention recall, and caption length ratio. Two human reviewers additionally score the generated captions with model identities hidden. Subject correctness, visual factuality, grounded informativeness, and language and format are each rated from 0 to 2, and their sum gives an overall score from 0 to 8. Every output receives one primary human score. A 200-output subset is also scored by both reviewers to measure agreement; these second scores are excluded from model means. Human-score intervals use 10,000 stratified paired-image bootstrap resamples; automatic-metric intervals use image-level bootstrap resampling.

\subsection{Main Classification and Retrieval Benchmark}
\label{sec:main_classification}

Mean Top-1 accuracy ranges from 49.08\% for SwinV2-Tiny~\citep{liu2022swinv2} to 84.39\% for MaxViT-Tiny~\citep{tu2022maxvit}. Mean Top-1 and Macro-F1 differ by less than one percentage point for 21 of the 23 models, showing close agreement between image-level and class-aware performance for most architectures. MaxViT-Tiny gives the highest mean on all four classification measures, whereas SwinV2-Tiny gives the lowest. Several convolutional and hybrid models occupy the upper part of the table. The self-supervised encoders also vary in performance: DINOv2 ViT-B/14~\citep{oquab2024dinov2} is near the top, while MAE ViT-B/16~\citep{he2022mae} is lower. Classification and retrieval rankings are not identical, with the largest differences appearing for class-macro mAP@R rather than Recall@1. MaxViT-Tiny leads mean R@1 and class-macro mAP@R, while RegNetY-8GF~\citep{radosavovic2020designing} leads mean R@5. RepViT-M1.0~\citep{wang2024repvit} has the lowest mean R@1, whereas SwinV2-Tiny has the lowest mean class-macro mAP@R.

Using the fixed seed-41 checkpoints, a scene-stratified analysis separates the test set into 949 controlled-scene and 3,009 intertidal-scene images. Controlled-scene Top-1 exceeds intertidal-scene Top-1 for 20 of the 23 models, and the largest absolute difference between the strata is 4.10 percentage points. Appendix Sec.~\ref{app:scene-recognition} reports these aggregate results and a taxon-matched recall analysis across three seeds for the 96 taxa represented in both strata. In the held-out benchmark, 20 of the 23 models have a Top-1 sample standard deviation below one percentage point; SwinV2-Tiny, iBOT ViT-B/16, and RepViT-M1.0 exceed that threshold. Dispersion is metric-dependent: the class-macro mAP@R sample standard deviation reaches 11.51 percentage points for iBOT ViT-B/16 and 7.40 points for PVTv2-B2. These standard deviations are descriptive summaries over three runs, not significance tests. In the separate taxonomy analysis, projecting the fixed held-out predictions to higher taxonomic ranks raises Top-1 for every model at both family and order levels. The mean gains over class-level accuracy are 4.05 and 8.27 percentage points, respectively. This pattern indicates that some fine-grained errors occur within the correct family or order. Appendix Sec.~\ref{app:taxonomy-recognition} provides the complete results.

\subsection{Captioning Benchmark}
\label{sec:captioning_results}

The captioning benchmark evaluates ten open image-captioning and vision-language models on the same 512 held-out images. The reference descriptions are the human expert annotations described above, prepared with draft assistance from GPT-5.6 Sol. Fig.~\ref{Fig:caption-metrics} reports the blinded human overall score together with four automatic metrics, each with a 95\% bootstrap interval. CIDEr measures TF-IDF-weighted n-gram agreement with the expert-finalized reference for each image, whereas SPICE compares semantic scene-graph content. CLIPScore measures image--text alignment, while BERTScore measures contextual textual similarity. Appendix Secs.~\ref{app:caption-human-evaluation} and~\ref{app:caption-metrics} provide the human dimension scores, agreement analysis, and complete automatic metric matrix.

The mean human score is 6.89/8 for Qwen3-VL-4B~\citep{bai2025qwen3vl} (95\% CI: 6.78--6.99), 6.82 for Qwen2.5-VL-3B~\citep{bai2025qwen25vl} (6.69--6.96), and 6.17 for LLaVA-1.5-7B~\citep{liu2024llava15} (6.00--6.34). LLaVA-Med 1.5 7B~\citep{li2023llavamed}, ViT-GPT2~\citep{nlpconnect2022vitgpt2,dosovitskiy2021image,radford2019gpt2}, and Phi-3.5 Vision~\citep{abdin2024phi3} receive lower overall means of 3.25, 2.57, and 2.51. Their mean language-and-format scores remain high (2.00, 1.85, and 1.89), while their subject-correctness means are 0.18, 0.09, and 0.03, respectively, separating well-formed output from correct visual identification. Qwen3-VL-4B also has the largest point estimate for each displayed automatic metric and leads BLEU-4, METEOR, ROUGE-L, and scientific-name token mention recall in the complete automatic evaluation. InternVL2.5-8B~\citep{chen2024internvl25} and Qwen2.5-VL-3B form the next group on most reference-based and image--text measures, although their order varies by metric. Appendix Sec.~\ref{app:caption-failure-cases} contrasts the expert-annotated references with exact model outputs in five qualitative failure cases.

\subsection{Targeted Synthetic Sensitivity Analyses}
\label{sec:synthetic_sensitivity}

We evaluate two sets of fixed seed-41 classifiers on the same 3,958 held-out images, applying every transformation independently and without parameter updates. The generic set comprises ResNet-50~\citep{he2016deep}, ConvNeXtV2-Tiny~\citep{woo2023convnextv2}, SwinV2-Tiny, MaxViT-Tiny, and DINOv2 ViT-B/14. It covers Gaussian noise, Gaussian blur, brightness, and contrast at three increasing levels. A separate marine-inspired set comprises ResNet-50, MobileNetV3-Large~\citep{howard2019searching}, EfficientFormerV2-S2~\citep{li2023efficientformerv2}, DINOv2 ViT-B/14, and RepViT-M1.0. Its two water-attenuation conditions follow the CADDY implementation of Jerlov II and III water types~\citep{gomezchavez2019caddy,gomezchavez2019underwaterdistortions}, while the marine-snow condition adapts the PHISWID synthesis to the RGB benchmark inputs~\citep{kaneko2026phiswid}. Figure~\ref{Fig:synthetic-sensitivity} summarizes Top-1 accuracy; Appendix Secs.~\ref{app:synthetic-sensitivity} and~\ref{app:marine-degradation} provide the exact settings and complete Top-1, Macro-F1, and Balanced Accuracy results. Accordingly, the analysis characterizes checkpoint-level rather than seed-averaged sensitivity.

\input{fig/figure4_layout}

In Figure~\ref{Fig:synthetic-sensitivity}(a), the generic five-model mean is 76.41\% on clean images. At level 3, it decreases to 35.37\% under Gaussian noise and 49.64\% under Gaussian blur, compared with 73.06\% under contrast change and 74.71\% under brightness change. Panel (b) expresses each marine-transformation result as a decrease from the corresponding model's clean Top-1. The five-model mean decreases are 23.59 percentage points for Jerlov II attenuation, 16.82 for marine snow, and 21.46 for Jerlov III attenuation. The model rows show a wider separation: RepViT-M1.0 loses 40.5 and 39.0 points under Jerlov II and III attenuation, whereas the three decreases for DINOv2 ViT-B/14 range from 7.8 to 13.0 points. Across these fixed transformations, sensitivity varies across appearance changes and evaluated fixed checkpoints.

\section{Conclusions and Future Work}


This paper introduces ShellfishNet, a benchmark of 40,130 images across 100 shellfish classes with provenance metadata and group-disjoint splits. Across representative visual models, image-level and class-aware classification measures generally agree, whereas classification and image-to-image retrieval produce different rankings. Post-hoc taxonomic projection shows that some exact-class errors fall within the correct family or order, indicating that species-level accuracy alone does not reveal the taxonomic proximity of all errors. Selected fixed checkpoints spanning different clean held-out accuracies can still exhibit distinct sensitivity to controlled generic and marine-inspired synthetic transformations. In the captioning study, human scores and automatic metrics assess different properties and yield metric-dependent model rankings. Overall, within ShellfishNet and the evaluated transformations, no single classification score captures all measured aspects of model performance, motivating task- and condition-aware comparison. We will release the dataset, splits, and evaluation protocols to support reproducible research on shellfish recognition and related marine vision tasks. Future work will extend ShellfishNet toward broader cross-dataset transfer evaluation, validation under real underwater conditions and on externally collected degraded imagery, broader geographic and temporal coverage, and larger-scale multi-reviewer evaluation of generated captions.

\clearpage

\subsection*{AI use statement}

Generative AI tools assisted with polishing the English manuscript. For the captioning subset, these tools also assisted with initial reference-caption drafts; a human domain expert led the annotation, inspected every image, revised the drafts, and determined the final wording. All AI-assisted text was verified against the relevant data sources or experimental outputs. The authors take full responsibility for the final manuscript, data, figures, code, and conclusions.

\subsection*{Ethics statement}

This work uses publicly accessible biodiversity and natural-history images rather than data collected from human participants or through animal experimentation. The released records retain source, attribution, and license information, and the included images are distributed under the licenses recorded for their sources, including CC0, CC BY, and CC BY-SA.

\subsection*{Reproducibility statement}

The main paper specifies the source platforms, inclusion and taxonomy checks, group-disjoint partitions, preprocessing, training, inference, and evaluation protocols. Appendix Sec.~\ref{app:benchmark-specification} documents the complete class inventory, partition statistics, and released record schema. Appendix Secs.~\ref{app:heldout-three-seed}, \ref{app:specialized-fgvc-baseline}, and~\ref{app:breadth-scan} provide the complete held-out results, the specialized fine-grained baseline, and the 80-backbone scan configuration, while Sec.~\ref{app:recognition-retrieval} reports complementary retrieval, few-shot prototype readout, taxonomy-aware, family-auxiliary, scene-stratified, shared-taxon transfer, and zero-shot analyses. Appendix Sec.~\ref{app:caption-evaluation} gives the model-specific captioning interfaces, complete metric matrix, and qualitative cases. Sec.~\ref{app:synthetic-analysis} gives the complete transformation-by-level results, and Sec.~\ref{app:additional-diagnostics} documents the marine-degradation and class-frequency diagnostics. The dataset manifest, provenance fields, experiment outputs, and accompanying implementation materials will be released with the paper to support independent reconstruction of the benchmark.

\input{references.bbl}
\clearpage
\appendix
\input{appendix}

\end{document}

%% file: math_commands.tex
\usepackage{amsmath,amsfonts,bm}

\def\eqref#1{equation~\ref{#1}}

\def\1{\bm{1}}

\DeclareMathAlphabet{\mathsfit}{\encodingdefault}{\sfdefault}{m}{sl}
\SetMathAlphabet{\mathsfit}{bold}{\encodingdefault}{\sfdefault}{bx}{n}



%% file: fig/figure1_layout.tex
\begin{figure}[t]
  \centering
  \hcenteredincludegraphics[width=\textwidth]{-0.3pt}{0.3pt}{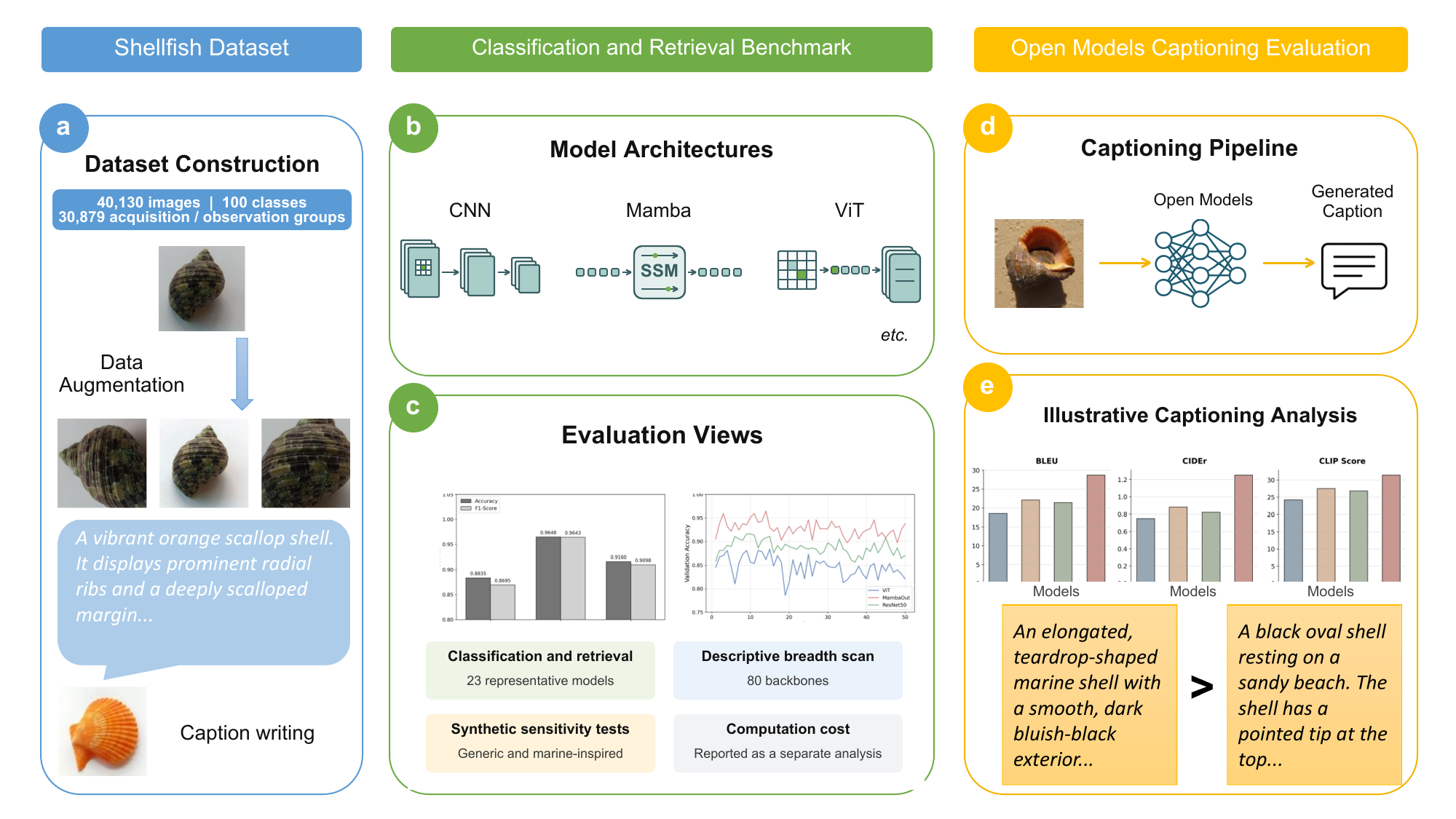}
  \caption{Overview of ShellfishNet from dataset construction to classification, retrieval, descriptive breadth and efficiency analyses, synthetic sensitivity tests, and open-model image captioning.}
  \label{Fig:figure1}
\end{figure}

%% file: fig/figure2_layout.tex
\begin{figure}[t]
    \centering
    \hcenteredincludegraphics[width=\textwidth]{9.8pt}{-9.8pt}{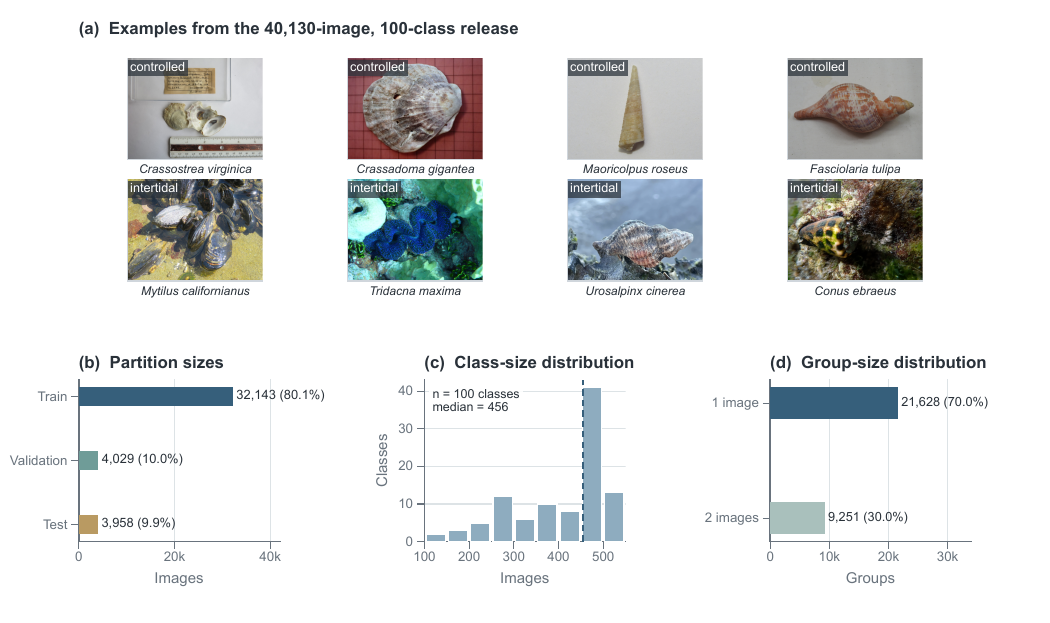}
    \caption{ShellfishNet dataset overview: (a) representative controlled and intertidal scenes; (b) partition sizes; (c) image counts across 100 classes; and (d) acquisition or observation group sizes.}
    \label{Fig:figure2}
\end{figure}

%% file: table/classification_main.tex
\begin{table}[t]
\centering
\small
\setlength{\tabcolsep}{2pt}
\caption{Held-out classification and retrieval means for 23 visual models. Values are percentages, and column-best means are bold. Sample standard deviations for all seven metrics and seed-specific Top-1 and class-macro mAP@R results appear in Appendix Sec.~\ref{app:heldout-three-seed}; mAP@R is computed as a class macro-average~\citep{musgrave2020metric}.}
\label{tab:classification-main}
\begin{tabular}{p{1.55in}rrrrrrr}
\toprule
Model & Top-1 & Top-5 & Macro-F1 & Balanced Acc. & R@1 & R@5 & mAP@R \\
\midrule
ResNet-50 & 79.82 & 93.92 & 79.61 & 79.47 & 77.75 & 87.70 & 46.61 \\
ResNet-101 & 80.99 & 94.72 & 80.87 & 80.54 & 79.55 & 88.25 & 50.81 \\
DenseNet-121 & 77.67 & 92.71 & 77.30 & 77.23 & 74.63 & 86.64 & 39.08 \\
EfficientNetV2-S & 82.35 & 94.60 & 82.31 & 82.16 & 82.06 & 87.70 & 69.74 \\
RegNetY-8GF & 83.33 & 95.54 & 83.10 & 83.03 & 81.83 & \textbf{89.96} & 51.09 \\
MobileNetV3-Large & 74.91 & 91.38 & 74.37 & 74.33 & 72.56 & 85.67 & 33.42 \\
ConvNeXt-Tiny & 82.10 & 95.16 & 82.01 & 81.75 & 81.79 & 87.19 & 67.44 \\
ConvNeXtV2-Tiny & 78.20 & 93.69 & 77.80 & 77.52 & 77.77 & 84.04 & 67.98 \\
ViT-B/16 & 75.59 & 91.44 & 75.61 & 75.29 & 75.08 & 80.67 & 65.77 \\
DeiT III-Small/16 & 74.76 & 92.13 & 74.68 & 74.19 & 73.27 & 81.88 & 52.53 \\
Swin-Tiny & 81.67 & 94.76 & 81.46 & 81.25 & 81.15 & 88.63 & 54.30 \\
SwinV2-Tiny & 49.08 & 76.58 & 45.93 & 46.97 & 46.36 & 72.48 & 4.04 \\
PVTv2-B2 & 79.63 & 94.00 & 79.32 & 79.08 & 79.32 & 86.33 & 56.62 \\
MaxViT-Tiny & \textbf{84.39} & \textbf{95.62} & \textbf{84.40} & \textbf{84.26} & \textbf{84.26} & 87.91 & \textbf{77.83} \\
EVA-02-Tiny & 70.87 & 88.79 & 70.44 & 70.17 & 69.18 & 77.72 & 56.22 \\
EfficientFormerV2-S2 & 77.40 & 92.77 & 76.88 & 76.86 & 77.43 & 84.50 & 60.20 \\
MAE ViT-B/16 & 72.97 & 90.98 & 72.85 & 72.35 & 71.46 & 80.76 & 49.03 \\
iBOT ViT-B/16 & 79.59 & 93.91 & 79.61 & 79.22 & 78.78 & 87.49 & 47.39 \\
DINOv2 ViT-B/14 & 82.71 & 94.87 & 82.75 & 82.31 & 82.95 & 87.79 & 69.66 \\
OpenCLIP ViT-B/16 & 75.12 & 91.19 & 74.93 & 74.56 & 74.20 & 80.95 & 65.04 \\
SigLIP2 Base patch16-224 & 78.58 & 92.73 & 78.36 & 77.94 & 78.15 & 84.46 & 64.37 \\
MambaOut-Tiny & 76.93 & 92.93 & 76.75 & 76.55 & 75.69 & 85.51 & 51.07 \\
RepViT-M1.0 & 51.37 & 77.24 & 50.30 & 50.41 & 34.73 & 59.68 & 4.21 \\
\bottomrule
\end{tabular}
\par\smallskip
\parbox{\linewidth}{\raggedright\footnotesize\textit{Model references.}
ResNet-50/101~\citep{he2016deep}; DenseNet-121~\citep{huang2017densely};
EfficientNetV2-S~\citep{tan2021efficientnetv2}; RegNetY-8GF~\citep{radosavovic2020designing};
MobileNetV3-Large~\citep{howard2019searching}; ConvNeXt-Tiny~\citep{liu2022convnet};
ConvNeXtV2-Tiny~\citep{woo2023convnextv2}; ViT-B/16~\citep{dosovitskiy2021image};
DeiT III-Small/16~\citep{touvron2022deit3}; Swin-Tiny~\citep{liu2021swin};
SwinV2-Tiny~\citep{liu2022swinv2}; PVTv2-B2~\citep{wang2022pvtv2};
MaxViT-Tiny~\citep{tu2022maxvit}; EVA-02-Tiny~\citep{fang2024eva02};
EfficientFormerV2-S2~\citep{li2023efficientformerv2}; MAE ViT-B/16~\citep{he2022mae};
iBOT ViT-B/16~\citep{zhou2022ibot}; DINOv2 ViT-B/14~\citep{oquab2024dinov2};
OpenCLIP ViT-B/16~\citep{cherti2023openclip}; SigLIP2 Base patch16-224~\citep{tschannen2025siglip2};
MambaOut-Tiny~\citep{yu2025mambaout}; RepViT-M1.0~\citep{wang2024repvit}.}
\end{table}

%% file: fig/figure3_layout.tex
\begin{figure}[t]
    \centering
    \hcenteredincludegraphics[width=0.98\linewidth]{-14.6pt}{14.6pt}{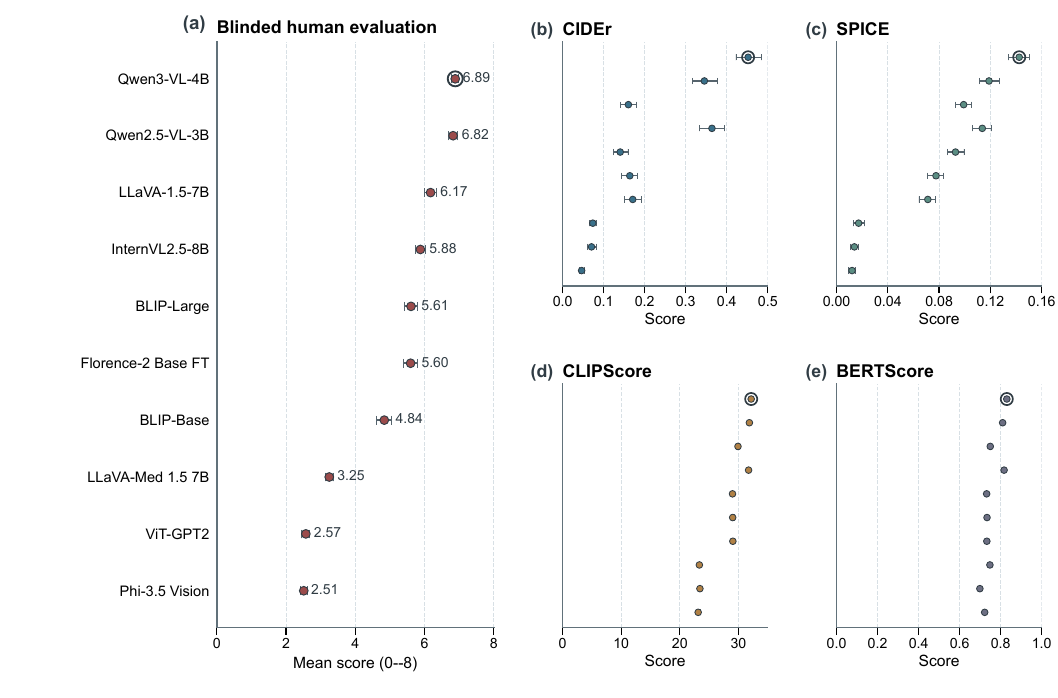}
    \caption{Human and automatic caption evaluation for ten open models on 512 images. Panel (a) shows blinded human overall scores, and panels (b)--(e) show CIDEr, SPICE, CLIPScore, and BERTScore. Horizontal intervals denote 95\% bootstrap confidence intervals; hollow rings mark panel-best point estimates.}
    \label{Fig:caption-metrics}
\end{figure}

%% file: fig/figure4_layout.tex
\begin{figure}[htbp]
  \centering
  \hcenteredincludegraphics[width=\linewidth]{-5.4pt}{5.4pt}{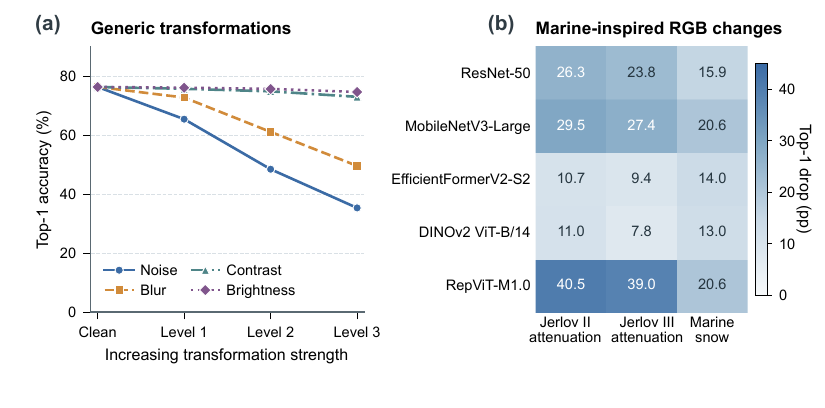}
  \caption{Top-1 sensitivity to controlled transformations on 3,958 held-out images. Panel (a) shows mean accuracy for five fixed checkpoints under Gaussian noise, Gaussian blur, brightness, and contrast at three levels. Panel (b) shows per-checkpoint decreases from clean accuracy for a separate five-checkpoint set under Jerlov II and III attenuation and marine snow; cell values are percentage-point decreases. Cell values are rounded to one decimal place; narrative averages use the underlying unrounded values.}
  \label{Fig:synthetic-sensitivity}
\end{figure}

%% file: appendix.tex
\begin{center}
{\Large\bfseries Appendix}
\end{center}

This appendix provides detailed dataset specifications and extended analyses of the classification, retrieval, captioning, and synthetic-sensitivity experiments. It includes the complete class inventory and partition statistics, three-seed held-out results, the 80-backbone descriptive scan, and complementary recognition and retrieval analyses. It also reports a specialized fine-grained baseline, a family-auxiliary comparison, and a shared-taxon transfer study. Further results cover the complete captioning evaluation, frozen zero-shot sensitivity checks, transformation-by-level results, marine-inspired RGB transformation tests, and a class-frequency analysis.

\section{Benchmark Specification and Data Partitions}
\label{app:benchmark-specification}

\subsection{Dataset Scope and Complete 100-Class Inventory}
\label{app:class-inventory}

The benchmark contains 40,130 images from 100 shellfish classes. The collection consolidates records from public biodiversity and media repositories. Sources include iNaturalist~\citep{van2018inaturalist}, GBIF~\citep{lane2007gbif}, Wikimedia Commons~\citep{vaidya2015dbpediacommons}, iDigBio~\citep{singer2018idigbio}, and the Encyclopedia of Life~\citep{parr2014eol}, among others. Table~\ref{tab:class-inventory} lists the complete taxonomic inventory and the image allocation for each class.

The 100 classes were selected from a broader pool of marine shellfish taxa through taxonomic, image-availability, and visual-discriminability screening. Each retained species required an accepted species-level name and consistent identifiers in WoRMS~\citep{worms2026}, GBIF, and iNaturalist, together with sufficient rights-cleared public imagery to support quality filtering and cross-source deduplication. Candidate taxa also had to exhibit diagnostic external features in ordinary photographs. We excluded taxa whose separation depended mainly on anatomy, microscopy, genetic evidence, or geographic context, as well as candidates with unresolved name collisions or severe visual overlap with an existing class. The resulting inventory spans Bivalvia, Gastropoda, and Polyplacophora and defines the fixed 100-class label space used in the classification and retrieval experiments.

\begin{small}
\begin{longtable}{>{\raggedright\arraybackslash}p{0.34\textwidth} >{\raggedright\arraybackslash}p{0.16\textwidth} r r r r}
\caption{Complete class inventory and image allocation. Scientific names follow the released taxonomy table.}\label{tab:class-inventory}\\
\toprule
Scientific name & Group & Total & Train & Val. & Test \\
\midrule
\endfirsthead
\multicolumn{6}{l}{\tablename~\thetable\ (continued)}\\
\toprule
Scientific name & Group & Total & Train & Val. & Test \\
\midrule
\endhead
\midrule
\multicolumn{6}{r}{Continued on next page}\\
\endfoot
\bottomrule
\endlastfoot
\textit{Crassostrea virginica} & Oyster & 500 & 400 & 51 & 49 \\
\textit{Ostrea edulis} & Oyster & 500 & 401 & 50 & 49 \\
\textit{Mytilus californianus} & Mussel & 500 & 401 & 50 & 49 \\
\textit{Geukensia demissa} & Mussel & 500 & 401 & 50 & 49 \\
\textit{Perna canaliculus} & Mussel & 500 & 400 & 51 & 49 \\
\textit{Argopecten irradians} & Scallop & 478 & 382 & 48 & 48 \\
\textit{Crassadoma gigantea} & Scallop & 486 & 390 & 49 & 47 \\
\textit{Tridacna maxima} & Giant\_Clam & 481 & 385 & 48 & 48 \\
\textit{Mercenaria mercenaria} & Venus\_Clam & 479 & 383 & 48 & 48 \\
\textit{Ruditapes philippinarum} & Venus\_Clam & 491 & 394 & 48 & 49 \\
\textit{Mya arenaria} & Soft\_Clam & 486 & 389 & 49 & 48 \\
\textit{Spisula solidissima} & Surf\_Clam & 482 & 387 & 48 & 47 \\
\textit{Donax variabilis} & Coquina & 465 & 372 & 47 & 46 \\
\textit{Cerastoderma edule} & Cockle & 482 & 386 & 49 & 47 \\
\textit{Dinocardium robustum} & Cockle & 455 & 364 & 46 & 45 \\
\textit{Ensis leei} & Razor\_Clam & 492 & 394 & 49 & 49 \\
\textit{Anomia simplex} & Jingle & 462 & 370 & 46 & 46 \\
\textit{Arca noae} & Ark\_Clam & 374 & 299 & 38 & 37 \\
\textit{Venus verrucosa} & Venus\_Clam & 426 & 341 & 43 & 42 \\
\textit{Cyrtopleura costata} & Angelwing & 451 & 361 & 45 & 45 \\
\textit{Nuttallia obscurata} & Tellin & 490 & 392 & 49 & 49 \\
\textit{Tivela stultorum} & Pismo\_Clam & 466 & 374 & 47 & 45 \\
\textit{Haliotis iris} & Abalone & 477 & 381 & 48 & 48 \\
\textit{Patella vulgata} & True\_Limpet & 476 & 381 & 48 & 47 \\
\textit{Lottia gigantea} & True\_Limpet & 489 & 391 & 49 & 49 \\
\textit{Diodora aspera} & Keyhole\_Limpet & 440 & 352 & 45 & 43 \\
\textit{Crepidula fornicata} & Slipper\_Snail & 488 & 390 & 49 & 49 \\
\textit{Littorina littorea} & Periwinkle & 477 & 381 & 48 & 48 \\
\textit{Littoraria irrorata} & Periwinkle & 352 & 283 & 34 & 35 \\
\textit{Bembicium nanum} & Periwinkle & 374 & 299 & 38 & 37 \\
\textit{Tegula funebralis} & Top\_Snail & 481 & 386 & 48 & 47 \\
\textit{Calliostoma ligatum} & Top\_Snail & 464 & 372 & 46 & 46 \\
\textit{Megastraea undosa} & Turban & 485 & 389 & 49 & 47 \\
\textit{Lunella undulata} & Turban & 416 & 334 & 42 & 40 \\
\textit{Neverita duplicata} & Moon\_Snail & 451 & 361 & 45 & 45 \\
\textit{Aliger gigas} & Conch & 445 & 357 & 45 & 43 \\
\textit{Strombus pugilis} & Conch & 425 & 340 & 43 & 42 \\
\textit{Californiconus californicus} & Cone & 379 & 303 & 38 & 38 \\
\textit{Hexaplex trunculus} & Murex & 464 & 371 & 47 & 46 \\
\textit{Dicathais orbita} & Murex & 473 & 379 & 47 & 47 \\
\textit{Nucella lapillus} & Dogwhelk & 470 & 376 & 47 & 47 \\
\textit{Buccinum undatum} & Whelk & 479 & 383 & 48 & 48 \\
\textit{Busycon carica} & Whelk & 469 & 375 & 47 & 47 \\
\textit{Busycotypus canaliculatus} & Whelk & 412 & 330 & 41 & 41 \\
\textit{Urosalpinx cinerea} & Oyster\_Drill & 318 & 254 & 32 & 32 \\
\textit{Ilyanassa obsoleta} & Mudsnail & 473 & 379 & 47 & 47 \\
\textit{Tritia reticulata} & Nassa & 456 & 366 & 46 & 44 \\
\textit{Oliva sayana} & Olive & 451 & 361 & 45 & 45 \\
\textit{Callianax biplicata} & Olive & 459 & 367 & 47 & 45 \\
\textit{Cabestana spengleri} & Triton & 458 & 366 & 46 & 46 \\
\textit{Mopalia muscosa} & Chiton & 492 & 394 & 49 & 49 \\
\textit{Cryptochiton stelleri} & Chiton & 487 & 389 & 49 & 49 \\
\textit{Tonicella lineata} & Chiton & 488 & 390 & 49 & 49 \\
\textit{Acanthopleura granulata} & Chiton & 483 & 387 & 49 & 47 \\
\textit{Sypharochiton pelliserpentis} & Chiton & 496 & 398 & 50 & 48 \\
\textit{Lunarca ovalis} & Ark\_Clam & 500 & 400 & 50 & 50 \\
\textit{Acanthocardia tuberculata} & Cockle & 500 & 400 & 50 & 50 \\
\textit{Raeta plicatella} & Duck\_Clam & 382 & 307 & 38 & 37 \\
\textit{Crassula aequilatera} & Triangle\_Clam & 371 & 298 & 37 & 36 \\
\textit{Atrina zelandica} & Pen\_Shell & 266 & 213 & 27 & 26 \\
\textit{Pecten maximus} & Scallop & 370 & 297 & 36 & 37 \\
\textit{Plicatula gibbosa} & Kitten\_Paw & 253 & 204 & 25 & 24 \\
\textit{Maoricolpus roseus} & Turret\_Snail & 500 & 401 & 50 & 49 \\
\textit{Nerita tessellata} & Nerite & 299 & 239 & 30 & 30 \\
\textit{Alcithoe arabica} & Volute & 266 & 213 & 28 & 25 \\
\textit{Katharina tunicata} & Chiton & 442 & 355 & 44 & 43 \\
\textit{Sinistrofulgur sinistrum} & Whelk & 434 & 347 & 45 & 42 \\
\textit{Megathura crenulata} & Keyhole\_Limpet & 332 & 266 & 34 & 32 \\
\textit{Scutus breviculus} & Shield\_Shell & 296 & 237 & 30 & 29 \\
\textit{Janthina janthina} & Violet\_Snail & 354 & 283 & 36 & 35 \\
\textit{Charonia lampas} & Triton & 271 & 218 & 27 & 26 \\
\textit{Cypraea tigris} & Cowrie & 267 & 213 & 27 & 27 \\
\textit{Melongena corona} & Crown\_Conch & 351 & 281 & 35 & 35 \\
\textit{Neoterebra dislocata} & Auger & 301 & 241 & 30 & 30 \\
\textit{Cittarium pica} & Top\_Snail & 276 & 221 & 28 & 27 \\
\textit{Conus ebraeus} & Cone & 239 & 191 & 25 & 23 \\
\textit{Cyphoma gibbosum} & Ovulid & 253 & 203 & 25 & 25 \\
\textit{Noetia ponderosa} & Ark\_Clam & 500 & 401 & 50 & 49 \\
\textit{Mimachlamys varia} & Scallop & 491 & 393 & 49 & 49 \\
\textit{Stramonita haemastoma} & Rock\_Shell & 463 & 372 & 46 & 45 \\
\textit{Monetaria annulus} & Cowrie & 317 & 253 & 32 & 32 \\
\textit{Phorcus lineatus} & Top\_Snail & 500 & 400 & 51 & 49 \\
\textit{Austrovenus stutchburyi} & Venus\_Clam & 500 & 400 & 51 & 49 \\
\textit{Spondylus gaederopus} & Thorny\_Oyster & 145 & 116 & 16 & 13 \\
\textit{Clinocardium nuttallii} & Cockle & 500 & 401 & 50 & 49 \\
\textit{Mactra stultorum} & Trough\_Clam & 500 & 401 & 50 & 49 \\
\textit{Ceratostoma foliatum} & Murex & 320 & 256 & 32 & 32 \\
\textit{Siliqua patula} & Razor\_Clam & 352 & 283 & 34 & 35 \\
\textit{Fissurella volcano} & Keyhole\_Limpet & 348 & 278 & 36 & 34 \\
\textit{Semicassis granulata} & Helmet & 201 & 162 & 19 & 20 \\
\textit{Monoplex parthenopeus} & Triton & 226 & 182 & 22 & 22 \\
\textit{Rapana venosa} & Rapa\_Whelk & 267 & 214 & 26 & 27 \\
\textit{Lutraria lutraria} & Otter\_Clam & 251 & 202 & 24 & 25 \\
\textit{Arctica islandica} & Quahog & 261 & 209 & 26 & 26 \\
\textit{Sinum perspectivum} & Baby\_Ear & 238 & 191 & 24 & 23 \\
\textit{Bulla gouldiana} & Bubble & 152 & 123 & 15 & 14 \\
\textit{Telescopium telescopium} & Mangrove\_Snail & 156 & 126 & 16 & 14 \\
\textit{Lambis lambis} & Spider\_Conch & 132 & 106 & 13 & 13 \\
\textit{Aporrhais pespelecani} & Pelican\_Foot & 205 & 164 & 21 & 20 \\
\textit{Fasciolaria tulipa} & Tulip\_Shell & 189 & 151 & 19 & 19 \\
\end{longtable}
\end{small}

\subsection{Released Record Schema}
\label{app:record-schema}

The released metadata supports taxonomic interpretation, source tracing, rights checking, scene-based analysis, and reconstruction of the benchmark partitions. Table~\ref{tab:record-schema} summarizes the corresponding field groups. The public release retains source, attribution, and license fields for each image.

\begin{small}
\begin{longtable}{>{\raggedright\arraybackslash}p{0.22\textwidth} >{\raggedright\arraybackslash}p{0.70\textwidth}}
\caption{Field groups in the released image-level metadata.}\label{tab:record-schema}\\
\toprule
Field group & Purpose \\
\midrule
\endfirsthead
\multicolumn{2}{l}{\tablename~\thetable\ (continued)}\\
\toprule
Field group & Purpose \\
\midrule
\endhead
\midrule
\multicolumn{2}{r}{Continued on next page}\\
\endfoot
\bottomrule
\endlastfoot
Taxonomic identity & Benchmark class and scientific name used to connect each image with the 100-class inventory. \\
Source provenance & Public source platform and source-record references retained for traceability. \\
Rights & License, attribution, and rights-verification information associated with each image. \\
File integrity & File checksum and image dimensions used to verify the released image representation. \\
Scene context & Controlled or intertidal scene label used in the scene-stratified analysis. \\
Grouping & Acquisition or observation grouping used to keep related images in one partition. \\
Partition & Training, validation, or test assignment used by all benchmark tasks. \\
\end{longtable}
\end{small}

Controlled and intertidal are the only scene labels used for quantitative stratification; the marine transformations are separate synthetic diagnostics. Ecological observation, aquaculture, and post-harvest use are application contexts rather than separate metadata strata. Class-label verification was conducted by one reviewer familiar with shellfish taxonomy; independent second-expert agreement was not measured.

The release audit connects the candidate-image deduplication records, the selection manifest, and the released asset table. Strict duplicate detection identified 216 clusters containing 442 of the 57,171 candidate images and 234 duplicate pairs. We removed 226 noncanonical cluster members. An additional 754 pairs covering 878 image paths were retained for review. After accounting for seven paths shared with the strict clusters, this review removed 871 further paths, yielding 1,097 unique exclusions. Every exclusion is absent from the selected and released records. All 40,130 selected records map one-to-one to the released manifest through the stored checksum and source-identity fields.

\begin{table}[htbp]
\centering
\caption{Deduplication, release-linkage, and partition audit.}
\label{tab:dedup-partition-audit}
\begin{small}
\begin{tabular}{@{}>{\raggedright\arraybackslash}p{0.56\textwidth} >{\raggedleft\arraybackslash}p{0.30\textwidth}@{}}
\toprule
Audit quantity & Result \\
\midrule
Candidate image inventory & 57,171 images \\
Strict duplicate clusters & 216 clusters; 442 members; 234 pairs \\
Noncanonical members removed from strict clusters & 226 paths \\
Additional reviewed paths removed & 871 paths \\
Excluded paths absent from the release & 1,097 / 1,097 \\
Selected-to-released record linkage & 40,130 / 40,130 \\
Acquisition or observation groups & 30,879 groups; at most 2 images per group \\
Groups spanning more than one split & 0 \\
\bottomrule
\end{tabular}
\end{small}
\end{table}

Partitioning assigns every image from a recorded acquisition or observation group to one split. The grouping key uses a session, observation, or specimen identifier when available; otherwise, it uses the source platform and source record. Group disjointness applies to this key; photographer identity, geographic location, and source platform are not partition keys. Cross-platform identity resolution is limited to the available identifiers.

\section{Extended Classification Results and Breadth Scan}
\label{app:classification-extensions}

\subsection{Training Configuration Registry}
\label{app:training-configuration-registry}

The training registry covers all combinations of 23 models and three seeds (41, 42, and 43), yielding 69 model--seed runs. The accompanying machine-readable registry records model construction, public initialization checkpoint and revision, input normalization, optimizer schedule, batch construction, and the 100-class output head. All runs use 100 epochs, cross-entropy without label smoothing, and per-epoch cosine annealing to zero. Each run has an effective batch size of 48, and validation Macro-F1 determines checkpoint selection. Training uses AMP FP16, whereas validation and test inference use FP32. Table~\ref{tab:training-configuration-registry} summarizes the optimizer assignments, and Table~\ref{tab:public-initialization-registry} lists the public model initializations.

The two optimizer recipes were selected using validation-only anchor experiments. MobileNetV3-Large served as the SGD anchor, and ViT-B/16 served as the AdamW anchor. For each anchor, six learning-rate and weight-decay candidates were evaluated for 30 epochs with seed 41. Validation Macro-F1 selected the recipe. The selected recipes were then assigned by the predefined model-family policy and held fixed for the formal 100-epoch runs. After the policy was fixed, there was no per-model search or test-set access. Table~\ref{tab:optimizer-anchor-search} reports the validation-only candidate grid. All reported benchmark values are outcomes of the evaluated fixed model configurations; they are not estimates of architecture-specific optima or capacity upper bounds.

\begin{table}[htbp]
\centering
\caption{Validation-only optimizer anchor search with seed 41 and a 30-epoch horizon. Validation Macro-F1 values are percentages; bold marks the selected recipe for each anchor.}
\label{tab:optimizer-anchor-search}
\begin{small}
\setlength{\tabcolsep}{5pt}
\begin{tabular}{@{}lrrr@{}}
\toprule
Anchor and optimizer & LR & WD & Val. Macro-F1 \\
\midrule
\multicolumn{4}{@{}l}{\textit{MobileNetV3-Large (SGD)}} \\
& $3\times10^{-3}$ & $10^{-4}$ & 72.46 \\
& $3\times10^{-3}$ & $5\times10^{-4}$ & 73.56 \\
& $10^{-2}$ & $10^{-4}$ & \textbf{74.97} \\
& $10^{-2}$ & $5\times10^{-4}$ & 73.97 \\
& $3\times10^{-2}$ & $10^{-4}$ & 72.86 \\
& $3\times10^{-2}$ & $5\times10^{-4}$ & 69.95 \\
\midrule
\multicolumn{4}{@{}l}{\textit{ViT-B/16 (AdamW)}} \\
& $10^{-5}$ & $0.01$ & 69.83 \\
& $10^{-5}$ & $0.05$ & 69.85 \\
& $3\times10^{-5}$ & $0.01$ & 73.27 \\
& $3\times10^{-5}$ & $0.05$ & \textbf{73.85} \\
& $10^{-4}$ & $0.01$ & 73.20 \\
& $10^{-4}$ & $0.05$ & 72.92 \\
\bottomrule
\end{tabular}
\end{small}
\end{table}

\begin{table}[H]
\centering
\caption{Training configuration registry for the 69 benchmark runs. Actual batch and gradient accumulation yield an effective batch size of 48 in every row.}
\label{tab:training-configuration-registry}
\begin{small}
\setlength{\tabcolsep}{3pt}
\begin{tabular}{@{}>{\raggedright\arraybackslash}p{0.46\textwidth} l r r r c@{}}
\toprule
Models & Optimizer & Runs & LR & WD & \shortstack{Batch $\times$\\accum.} \\
\midrule
ResNet-50; ResNet-101; DenseNet-121; EfficientNetV2-S; RegNetY-8GF; MobileNetV3-Large; RepViT-M1.0 & SGD ($m=0.9$) & 21 & $10^{-2}$ & $10^{-4}$ & $48\times1$ \\
ConvNeXt-Tiny; ConvNeXtV2-Tiny; ViT-B/16; DeiT III-Small/16; Swin-Tiny; SwinV2-Tiny; PVTv2-B2; MaxViT-Tiny; EVA-02-Tiny; EfficientFormerV2-S2; MAE ViT-B/16; iBOT ViT-B/16; DINOv2 ViT-B/14; OpenCLIP ViT-B/16; SigLIP2 Base patch16-224 & AdamW & 45 & $3\times10^{-5}$ & $0.05$ & $48\times1$ \\
MambaOut-Tiny & AdamW & 3 & $3\times10^{-5}$ & $0.05$ & $24\times2$ \\
\bottomrule
\end{tabular}
\par\smallskip
\parbox{\linewidth}{\raggedright\footnotesize\textit{Model references.}
ResNet-50/101~\citep{he2016deep}; DenseNet-121~\citep{huang2017densely};
EfficientNetV2-S~\citep{tan2021efficientnetv2}; RegNetY-8GF~\citep{radosavovic2020designing};
MobileNetV3-Large~\citep{howard2019searching}; ConvNeXt-Tiny~\citep{liu2022convnet};
ConvNeXtV2-Tiny~\citep{woo2023convnextv2}; ViT-B/16~\citep{dosovitskiy2021image};
DeiT III-Small/16~\citep{touvron2022deit3}; Swin-Tiny~\citep{liu2021swin};
SwinV2-Tiny~\citep{liu2022swinv2}; PVTv2-B2~\citep{wang2022pvtv2};
MaxViT-Tiny~\citep{tu2022maxvit}; EVA-02-Tiny~\citep{fang2024eva02};
EfficientFormerV2-S2~\citep{li2023efficientformerv2}; MAE ViT-B/16~\citep{he2022mae};
iBOT ViT-B/16~\citep{zhou2022ibot}; DINOv2 ViT-B/14~\citep{oquab2024dinov2};
OpenCLIP ViT-B/16~\citep{cherti2023openclip}; SigLIP2 Base patch16-224~\citep{tschannen2025siglip2};
MambaOut-Tiny~\citep{yu2025mambaout}; RepViT-M1.0~\citep{wang2024repvit}.}
\end{small}
\end{table}

\begin{small}
\setlength{\tabcolsep}{3pt}
\begin{longtable}{@{}>{\raggedright\arraybackslash}p{0.19\textwidth} >{\raggedright\arraybackslash}p{0.28\textwidth} >{\raggedright\arraybackslash}p{0.378\textwidth} >{\centering\arraybackslash}p{0.082\textwidth}@{}}
\caption{Public implementation identifiers, initialization revisions, and retrieval dimensions for the 23-model benchmark.}\label{tab:public-initialization-registry}\\
\toprule
Model & Implementation identifier & Public initialization revision & Retrieval dim. \\
\midrule
\endfirsthead
\multicolumn{4}{l}{\tablename~\thetable\ (continued)}\\
\toprule
Model & Implementation identifier & Public initialization revision & Retrieval dim. \\
\midrule
\endhead
\midrule
\multicolumn{4}{r}{Continued on next page}\\
\endfoot
\bottomrule
\endlastfoot
ResNet-50 & \nolinkurl{torchvision.models.resnet50} & \nolinkurl{torchvision==0.16.2:ResNet50_Weights.IMAGENET1K_V2} & 2048 \\
ResNet-101 & \nolinkurl{torchvision.models.resnet101} & \nolinkurl{torchvision==0.16.2:ResNet101_Weights.IMAGENET1K_V2} & 2048 \\
DenseNet-121 & \nolinkurl{torchvision.models.densenet121} & \nolinkurl{torchvision==0.16.2:DenseNet121_Weights.IMAGENET1K_V1} & 1024 \\
EfficientNetV2-S & \nolinkurl{torchvision.models.efficientnet_v2_s} & \nolinkurl{torchvision==0.16.2:EfficientNet_V2_S_Weights.IMAGENET1K_V1} & 1280 \\
RegNetY-8GF & \nolinkurl{torchvision.models.regnet_y_8gf} & \nolinkurl{torchvision==0.16.2:RegNet_Y_8GF_Weights.IMAGENET1K_V2} & 2016 \\
MobileNetV3-Large & \nolinkurl{torchvision.models.mobilenet_v3_large} & \nolinkurl{torchvision==0.16.2:MobileNet_V3_Large_Weights.IMAGENET1K_V2} & 1280 \\
ConvNeXt-Tiny & \nolinkurl{torchvision.models.convnext_tiny} & \nolinkurl{torchvision==0.16.2:ConvNeXt_Tiny_Weights.IMAGENET1K_V1} & 768 \\
ConvNeXtV2-Tiny & \nolinkurl{convnextv2_tiny.fcmae_ft_in1k} & \nolinkurl{508caf1bc24f} & 768 \\
ViT-B/16 & \nolinkurl{torchvision.models.vit_b_16} & \nolinkurl{torchvision==0.16.2:ViT_B_16_Weights.IMAGENET1K_V1} & 768 \\
DeiT III-Small/16 & \nolinkurl{deit3_small_patch16_224.fb_in1k} & \nolinkurl{f3f500f06a4e} & 384 \\
Swin-Tiny & \nolinkurl{torchvision.models.swin_t} & \nolinkurl{torchvision==0.16.2:Swin_T_Weights.IMAGENET1K_V1} & 768 \\
SwinV2-Tiny & \nolinkurl{torchvision.models.swin_v2_t} & \nolinkurl{torchvision==0.16.2:Swin_V2_T_Weights.IMAGENET1K_V1} & 768 \\
PVTv2-B2 & \nolinkurl{pvt_v2_b2.in1k} & \nolinkurl{8dca9e485d6a} & 512 \\
MaxViT-Tiny & \nolinkurl{torchvision.models.maxvit_t} & \nolinkurl{torchvision==0.16.2:MaxVit_T_Weights.IMAGENET1K_V1} & 512 \\
EVA-02-Tiny & \nolinkurl{eva02_tiny_patch14_224.mim_in22k} & \nolinkurl{e0a5e8afdf15} & 192 \\
EfficientFormerV2-S2 & \nolinkurl{efficientformerv2_s2.snap_dist_in1k} & \nolinkurl{1c56a7635500} & 288 \\
MAE ViT-B/16 & \nolinkurl{vit_base_patch16_224.mae} & \nolinkurl{a64dffd137c0} & 768 \\
iBOT ViT-B/16 & \nolinkurl{vit_base_patch16_224.ibot_teacher} & \nolinkurl{da316d82636a} & 768 \\
DINOv2 ViT-B/14 & \nolinkurl{vit_base_patch14_dinov2.lvd142m} & \nolinkurl{4685c99dabff} & 768 \\
OpenCLIP ViT-B/16 & \nolinkurl{open-clip-torch==2.24.0}\linebreak\nolinkurl{ViT-B-16::openai} & \nolinkurl{openai:5806e77cd80f} & 512 \\
SigLIP2 Base patch16-224 & \texttt{google/siglip2-base-}\linebreak\texttt{patch16-224} & \nolinkurl{75de2d55ec2d} & 768 \\
MambaOut-Tiny & \nolinkurl{mambaout_tiny.in1k} & \nolinkurl{d080941663a6} & 2304 \\
RepViT-M1.0 & \nolinkurl{repvit_m1_0} & \nolinkurl{v1.0:repvit_m1_0_distill_450e_timm.pth} & 448 \\
\end{longtable}
\end{small}

Git-style hashes in Table~\ref{tab:public-initialization-registry} are abbreviated to 12 characters; the machine-readable registry retains the complete revisions. Complete image-level manifests are unavailable for several public pretraining corpora, leaving pretraining overlap unresolved for some encoders. The reported results describe performance on the fixed ShellfishNet evaluation. Section~\ref{app:zero-shot-classification} gives a provider-level analysis for BioCLIP~2.

\subsection{Three-Seed Held-Out Results for the 23-Model Benchmark}
\label{app:heldout-three-seed}

\begin{figure}[H]
  \centering
  \hcenteredincludegraphics[width=\linewidth]{0.1pt}{-0.1pt}{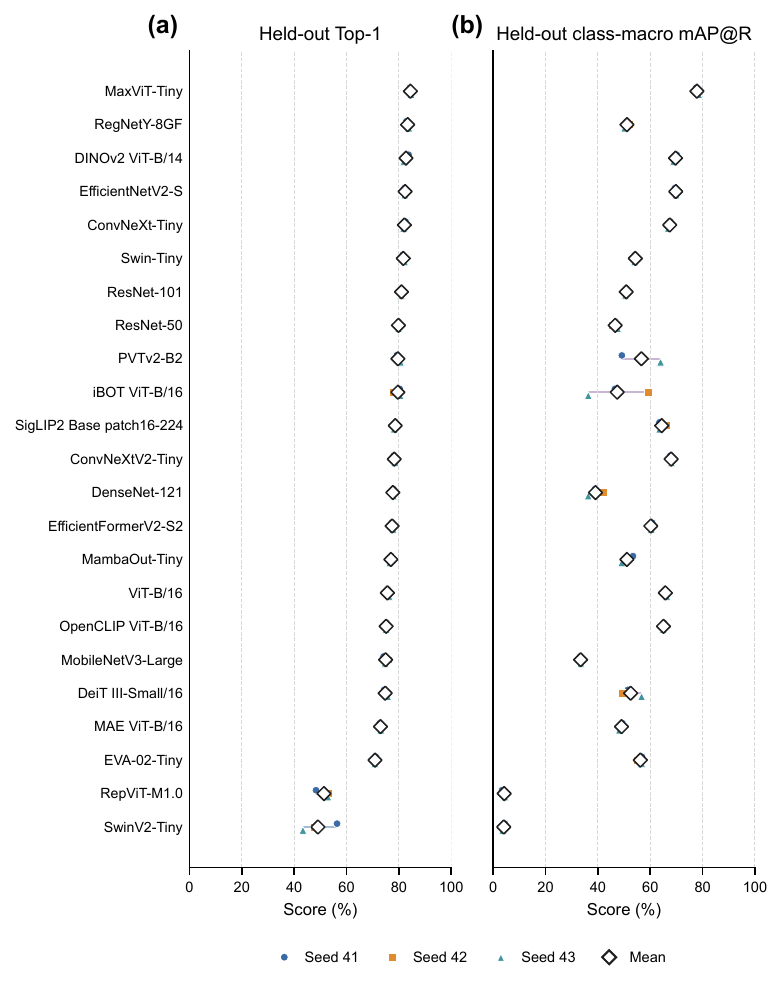}
  \caption{Seed-specific held-out Top-1 and class-macro mAP@R results for 23 models. Circles, squares, and triangles denote seeds 41, 42, and 43; horizontal segments span the observed minimum and maximum, and hollow diamonds mark arithmetic means. Rows are ordered by mean Top-1. Both panels use the full 0--100 percentage scale.}
  \label{fig:seed-repeatability}
\end{figure}

Each of the 23 models was trained with random seeds 41, 42, and 43 under the same model-specific configuration, yielding 69 runs. For each model and metric, Tables~\ref{tab:heldout-classification-23x3} and~\ref{tab:heldout-retrieval-23x3} report the arithmetic mean and sample standard deviation across the three held-out evaluations. Figure~\ref{fig:seed-repeatability} retains the seed-specific Top-1 and class-macro mAP@R values so that classification and retrieval dispersion can be inspected directly.

\begin{table}[H]
\centering
\small
\setlength{\tabcolsep}{3pt}
\caption{Complete three-seed held-out classification results. Values are mean $\pm$ sample SD over seeds 41, 42, and 43, in percentage points; column-best means are bold.}
\label{tab:heldout-classification-23x3}
\begin{tabular}{p{1.62in}cccc}
\toprule
Model & Top-1 & Top-5 & Macro-F1 & Balanced Acc. \\
\midrule
ResNet-50 & 79.82$\pm$0.05 & 93.92$\pm$0.08 & 79.61$\pm$0.12 & 79.47$\pm$0.14 \\
ResNet-101 & 80.99$\pm$0.10 & 94.72$\pm$0.20 & 80.87$\pm$0.12 & 80.54$\pm$0.10 \\
DenseNet-121 & 77.67$\pm$0.47 & 92.71$\pm$0.28 & 77.30$\pm$0.26 & 77.23$\pm$0.27 \\
EfficientNetV2-S & 82.35$\pm$0.29 & 94.60$\pm$0.05 & 82.31$\pm$0.40 & 82.16$\pm$0.42 \\
RegNetY-8GF & 83.33$\pm$0.47 & 95.54$\pm$0.15 & 83.10$\pm$0.50 & 83.03$\pm$0.47 \\
MobileNetV3-Large & 74.91$\pm$0.87 & 91.38$\pm$0.26 & 74.37$\pm$0.91 & 74.33$\pm$0.99 \\
ConvNeXt-Tiny & 82.10$\pm$0.42 & 95.16$\pm$0.19 & 82.01$\pm$0.41 & 81.75$\pm$0.31 \\
ConvNeXtV2-Tiny & 78.20$\pm$0.32 & 93.69$\pm$0.13 & 77.80$\pm$0.50 & 77.52$\pm$0.46 \\
ViT-B/16 & 75.59$\pm$0.73 & 91.44$\pm$0.34 & 75.61$\pm$0.80 & 75.29$\pm$0.91 \\
DeiT III-Small/16 & 74.76$\pm$0.81 & 92.13$\pm$0.10 & 74.68$\pm$0.92 & 74.19$\pm$0.94 \\
Swin-Tiny & 81.67$\pm$0.43 & 94.76$\pm$0.27 & 81.46$\pm$0.57 & 81.25$\pm$0.62 \\
SwinV2-Tiny & 49.08$\pm$6.64 & 76.58$\pm$5.28 & 45.93$\pm$6.82 & 46.97$\pm$7.05 \\
PVTv2-B2 & 79.63$\pm$0.85 & 94.00$\pm$0.47 & 79.32$\pm$1.04 & 79.08$\pm$1.03 \\
MaxViT-Tiny & \textbf{84.39$\pm$0.22} & \textbf{95.62$\pm$0.11} & \textbf{84.40$\pm$0.29} & \textbf{84.26$\pm$0.44} \\
EVA-02-Tiny & 70.87$\pm$0.28 & 88.79$\pm$0.16 & 70.44$\pm$0.25 & 70.17$\pm$0.29 \\
EfficientFormerV2-S2 & 77.40$\pm$0.31 & 92.77$\pm$0.24 & 76.88$\pm$0.38 & 76.86$\pm$0.31 \\
MAE ViT-B/16 & 72.97$\pm$0.33 & 90.98$\pm$0.28 & 72.85$\pm$0.21 & 72.35$\pm$0.30 \\
iBOT ViT-B/16 & 79.59$\pm$1.40 & 93.91$\pm$0.66 & 79.61$\pm$1.55 & 79.22$\pm$1.43 \\
DINOv2 ViT-B/14 & 82.71$\pm$0.94 & 94.87$\pm$0.09 & 82.75$\pm$1.00 & 82.31$\pm$1.01 \\
OpenCLIP ViT-B/16 & 75.12$\pm$0.13 & 91.19$\pm$0.03 & 74.93$\pm$0.24 & 74.56$\pm$0.11 \\
SigLIP2 Base patch16-224 & 78.58$\pm$0.32 & 92.73$\pm$0.12 & 78.36$\pm$0.67 & 77.94$\pm$0.71 \\
MambaOut-Tiny & 76.93$\pm$0.38 & 92.93$\pm$0.23 & 76.75$\pm$0.42 & 76.55$\pm$0.37 \\
RepViT-M1.0 & 51.37$\pm$2.63 & 77.24$\pm$1.85 & 50.30$\pm$2.87 & 50.41$\pm$2.96 \\
\bottomrule
\end{tabular}
\par\smallskip
\parbox{\linewidth}{\raggedright\footnotesize\textit{Model references.}
ResNet-50/101~\citep{he2016deep}; DenseNet-121~\citep{huang2017densely};
EfficientNetV2-S~\citep{tan2021efficientnetv2}; RegNetY-8GF~\citep{radosavovic2020designing};
MobileNetV3-Large~\citep{howard2019searching}; ConvNeXt-Tiny~\citep{liu2022convnet};
ConvNeXtV2-Tiny~\citep{woo2023convnextv2}; ViT-B/16~\citep{dosovitskiy2021image};
DeiT III-Small/16~\citep{touvron2022deit3}; Swin-Tiny~\citep{liu2021swin};
SwinV2-Tiny~\citep{liu2022swinv2}; PVTv2-B2~\citep{wang2022pvtv2};
MaxViT-Tiny~\citep{tu2022maxvit}; EVA-02-Tiny~\citep{fang2024eva02};
EfficientFormerV2-S2~\citep{li2023efficientformerv2}; MAE ViT-B/16~\citep{he2022mae};
iBOT ViT-B/16~\citep{zhou2022ibot}; DINOv2 ViT-B/14~\citep{oquab2024dinov2};
OpenCLIP ViT-B/16~\citep{cherti2023openclip}; SigLIP2 Base patch16-224~\citep{tschannen2025siglip2};
MambaOut-Tiny~\citep{yu2025mambaout}; RepViT-M1.0~\citep{wang2024repvit}.}
\end{table}

\begin{table}[H]
\centering
\small
\setlength{\tabcolsep}{3pt}
\caption{Complete three-seed held-out retrieval results. Values are mean $\pm$ sample SD over seeds 41, 42, and 43, in percentage points; column-best means are bold. mAP@R denotes the class-macro average~\citep{musgrave2020metric}.}
\label{tab:heldout-retrieval-23x3}
\begin{tabular}{p{1.62in}ccc}
\toprule
Model & R@1 & R@5 & mAP@R \\
\midrule
ResNet-50 & 77.75$\pm$0.23 & 87.70$\pm$0.42 & 46.61$\pm$0.93 \\
ResNet-101 & 79.55$\pm$0.40 & 88.25$\pm$0.24 & 50.81$\pm$0.29 \\
DenseNet-121 & 74.63$\pm$0.25 & 86.64$\pm$0.98 & 39.08$\pm$2.89 \\
EfficientNetV2-S & 82.06$\pm$0.37 & 87.70$\pm$0.03 & 69.74$\pm$0.35 \\
RegNetY-8GF & 81.83$\pm$0.42 & \textbf{89.96$\pm$0.15} & 51.09$\pm$0.87 \\
MobileNetV3-Large & 72.56$\pm$0.78 & 85.67$\pm$0.54 & 33.42$\pm$0.13 \\
ConvNeXt-Tiny & 81.79$\pm$0.28 & 87.19$\pm$0.24 & 67.44$\pm$0.51 \\
ConvNeXtV2-Tiny & 77.77$\pm$0.40 & 84.04$\pm$0.63 & 67.98$\pm$0.30 \\
ViT-B/16 & 75.08$\pm$0.88 & 80.67$\pm$0.78 & 65.77$\pm$0.56 \\
DeiT III-Small/16 & 73.27$\pm$0.58 & 81.88$\pm$0.27 & 52.53$\pm$3.73 \\
Swin-Tiny & 81.15$\pm$0.59 & 88.63$\pm$0.28 & 54.30$\pm$0.35 \\
SwinV2-Tiny & 46.36$\pm$1.68 & 72.48$\pm$1.73 & 4.04$\pm$0.44 \\
PVTv2-B2 & 79.32$\pm$1.31 & 86.33$\pm$0.65 & 56.62$\pm$7.40 \\
MaxViT-Tiny & \textbf{84.26$\pm$0.66} & 87.91$\pm$0.31 & \textbf{77.83$\pm$0.59} \\
EVA-02-Tiny & 69.18$\pm$0.34 & 77.72$\pm$0.22 & 56.22$\pm$1.02 \\
EfficientFormerV2-S2 & 77.43$\pm$0.86 & 84.50$\pm$0.35 & 60.20$\pm$0.63 \\
MAE ViT-B/16 & 71.46$\pm$0.17 & 80.76$\pm$0.43 & 49.03$\pm$0.74 \\
iBOT ViT-B/16 & 78.78$\pm$0.91 & 87.49$\pm$2.41 & 47.39$\pm$11.51 \\
DINOv2 ViT-B/14 & 82.95$\pm$0.79 & 87.79$\pm$0.54 & 69.66$\pm$0.63 \\
OpenCLIP ViT-B/16 & 74.20$\pm$0.33 & 80.95$\pm$0.32 & 65.04$\pm$0.16 \\
SigLIP2 Base patch16-224 & 78.15$\pm$0.25 & 84.46$\pm$0.23 & 64.37$\pm$1.43 \\
MambaOut-Tiny & 75.69$\pm$0.45 & 85.51$\pm$0.21 & 51.07$\pm$2.18 \\
RepViT-M1.0 & 34.73$\pm$2.30 & 59.68$\pm$1.92 & 4.21$\pm$0.76 \\
\bottomrule
\end{tabular}
\par\smallskip
\parbox{\linewidth}{\raggedright\footnotesize\textit{Model references.}
ResNet-50/101~\citep{he2016deep}; DenseNet-121~\citep{huang2017densely};
EfficientNetV2-S~\citep{tan2021efficientnetv2}; RegNetY-8GF~\citep{radosavovic2020designing};
MobileNetV3-Large~\citep{howard2019searching}; ConvNeXt-Tiny~\citep{liu2022convnet};
ConvNeXtV2-Tiny~\citep{woo2023convnextv2}; ViT-B/16~\citep{dosovitskiy2021image};
DeiT III-Small/16~\citep{touvron2022deit3}; Swin-Tiny~\citep{liu2021swin};
SwinV2-Tiny~\citep{liu2022swinv2}; PVTv2-B2~\citep{wang2022pvtv2};
MaxViT-Tiny~\citep{tu2022maxvit}; EVA-02-Tiny~\citep{fang2024eva02};
EfficientFormerV2-S2~\citep{li2023efficientformerv2}; MAE ViT-B/16~\citep{he2022mae};
iBOT ViT-B/16~\citep{zhou2022ibot}; DINOv2 ViT-B/14~\citep{oquab2024dinov2};
OpenCLIP ViT-B/16~\citep{cherti2023openclip}; SigLIP2 Base patch16-224~\citep{tschannen2025siglip2};
MambaOut-Tiny~\citep{yu2025mambaout}; RepViT-M1.0~\citep{wang2024repvit}.}
\end{table}

\subsection{Specialized Fine-Grained Baseline}
\label{app:specialized-fgvc-baseline}

We evaluate HERBS with a Swin-L/384 backbone as a specialized fine-grained baseline. The model uses the released selector, feature-pyramid, and combiner design. It is trained for 80 epochs with SGD-Nesterov, an effective batch size of 32, a maximum learning rate of $5\times10^{-4}$, and weight decay of $3\times10^{-4}$. Validation Macro-F1 selects the epoch-56 checkpoint, which is evaluated on the held-out test partition in FP32. Table~\ref{tab:herbs-swinl384} reports 91.66\% Top-1 accuracy and 91.93\% Macro-F1. This single-run result remains separate from the three-run, 224-pixel benchmark ranking.

\begin{table}[H]
\centering
\small
\setlength{\tabcolsep}{5pt}
\caption{Held-out classification results for the specialized fine-grained baseline. All metrics are percentages from one run.}
\label{tab:herbs-swinl384}
\begin{tabular}{lrrrrrrr}
\toprule
Model & Runs & Input & Epochs & Top-1 & Top-5 & Macro-F1 & Bal.\ acc. \\
\midrule
HERBS (Swin-L/384) & 1 & 384 & 80 & 91.66 & 98.23 & 91.93 & 91.71 \\
\bottomrule
\end{tabular}
\par\smallskip
\parbox{\linewidth}{\raggedright\footnotesize\textit{Model references.}
HERBS~\citep{chou2023herbs}; released implementation~\citep{chou2023herbssoftware};
Swin Transformer~\citep{liu2021swin}.}
\end{table}

\subsection{Complete 80-Backbone Descriptive Scan}
\label{app:breadth-scan}

\begin{figure}[htbp]
  \centering
  \hcenteredincludegraphics[width=0.98\linewidth]{-8.6pt}{8.6pt}{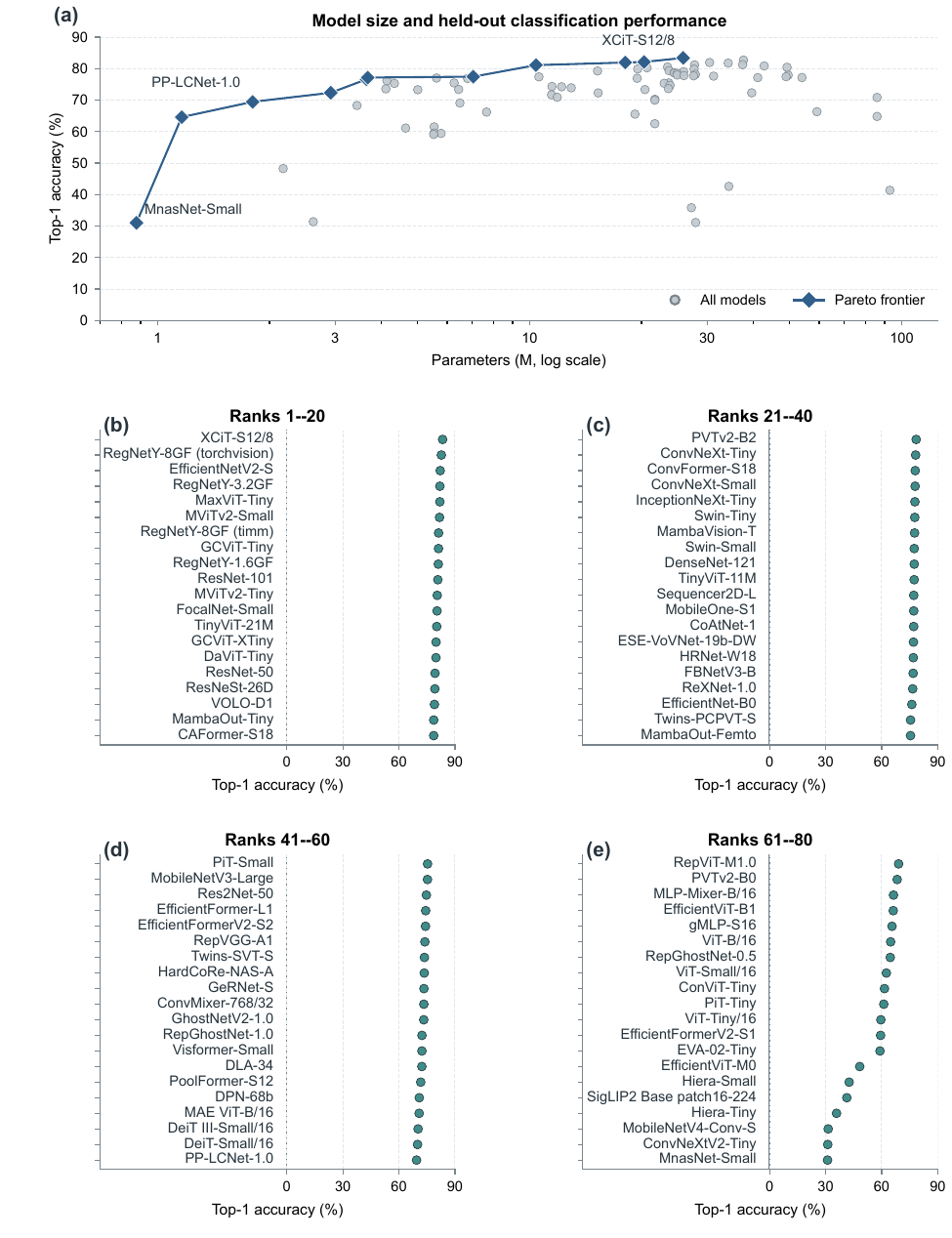}
  \caption{Complete 80-backbone descriptive scan. Panel (a) relates parameter count to held-out Top-1 accuracy and marks the Pareto frontier. Panels (b)--(e) show the observed Top-1 ordering for all 80 models in groups of 20. Each model uses one seed and a 30-epoch training horizon.}
  \label{fig:breadth-overview}
\end{figure}

The breadth scan evaluates 80 backbones on the same 100-class task. All runs use seed 41, 30 epochs, $224\times224$ inputs, ImageNet normalization, and an effective batch size of 48. Optimization uses SGD with an initial learning rate of 0.01, weight decay of 0.0005, and cosine annealing. Figure~\ref{fig:breadth-overview} summarizes the model-size tradeoff and displays the complete Top-1 ordering across all 80 backbones. Validation Macro-F1 selects the checkpoint within the 30-epoch horizon. The accompanying machine-readable registry reports each backbone's implementation identifier, pretrained initialization, micro-batch size, gradient-accumulation setting, and selected epoch. The scan describes architectural breadth rather than seed-stable rankings. Parameter counts support direct cross-model comparison, while timing and throughput retain their run-specific hardware context in Table~\ref{tab:breadth-efficiency}.

\begin{small}
\begin{longtable}{@{}>{\raggedright\arraybackslash}p{0.42\textwidth} >{\centering\arraybackslash}p{0.14\textwidth} >{\centering\arraybackslash}p{0.14\textwidth} >{\centering\arraybackslash}p{0.14\textwidth}@{}}
\caption{Computational measurements for the 80-backbone descriptive scan. Paired entries report parameter count/peak memory, training throughput/duration, and test throughput/duration; timing and throughput reflect the hardware used for each run.}\label{tab:breadth-efficiency}\\
\toprule
Model & \shortstack{Parameters\\Peak memory\\(M / MiB)} & \shortstack{Training\\rate / time\\(images/s /\\min)} & \shortstack{Testing\\rate / time\\(images/s /\\s)} \\
\midrule
\endfirsthead
\multicolumn{4}{l}{\tablename~\thetable\ (continued)}\\
\toprule
Model & \shortstack{Parameters\\Peak memory\\(M / MiB)} & \shortstack{Training\\rate / time\\(images/s /\\min)} & \shortstack{Testing\\rate / time\\(images/s /\\s)} \\
\midrule
\endhead
\midrule
\multicolumn{4}{r}{Continued on next page}\\
\endfoot
\bottomrule
\endlastfoot
DaViT-Tiny & 27.67 / 2856 & 244.9 / 65.6 & 555.2 / 7.13 \\
DLA-34 & 15.28 / 1482 & 239.8 / 67.0 & 575.1 / 6.88 \\
HRNet-W18 & 19.45 / 2506 & 138.7 / 115.9 & 520.8 / 7.60 \\
ConvMixer-768/32 & 20.42 / 5908 & 62.8 / 256.0 & 124.3 / 31.84 \\
ResNeSt-26D & 15.23 / 2716 & 235.2 / 68.3 & 553.7 / 7.15 \\
FBNetV3-B & 6.81 / 2830 & 244.4 / 65.8 & 531.0 / 7.45 \\
ConvNeXt-Small & 49.53 / 2144 & 275.8 / 58.3 & 686.5 / 5.77 \\
Sequencer2D-L & 53.95 / 3268 & 110.0 / 146.2 & 377.3 / 10.49 \\
XCiT-S12/8 & 25.87 / 6698 & 101.0 / 159.1 & 179.2 / 22.09 \\
Res2Net-50 & 23.86 / 2366 & 244.9 / 65.6 & 559.9 / 7.07 \\
EfficientNet-B0 & 4.14 / 3090 & 242.8 / 66.2 & 560.7 / 7.06 \\
PoolFormer-S12 & 11.45 / 2624 & 236.0 / 68.1 & 540.7 / 7.32 \\
GCViT-XTiny & 19.52 / 3322 & 198.0 / 81.2 & 514.5 / 7.69 \\
PiT-Tiny & 4.64 / 848 & 243.5 / 66.0 & 574.4 / 6.89 \\
PiT-Small & 23.00 / 1292 & 235.7 / 68.2 & 545.3 / 7.26 \\
RegNetY-1.6GF & 10.40 / 2314 & 233.8 / 68.7 & 548.8 / 7.21 \\
gMLP-S16 & 19.19 / 2870 & 245.8 / 65.4 & 554.9 / 7.13 \\
RegNetY-8GF (torchvision) & 37.36 / 2126 & 314.1 / 51.2 & 658.2 / 6.01 \\
ReXNet-1.0 & 3.64 / 3264 & 241.2 / 66.6 & 555.8 / 7.12 \\
MLP-Mixer-B/16 & 59.19 / 1862 & 316.1 / 50.8 & 666.1 / 5.94 \\
GhostNetV2-1.0 & 5.00 / 1982 & 244.7 / 65.7 & 572.2 / 6.92 \\
RepGhostNet-1.0 & 2.92 / 1410 & 231.9 / 69.3 & 545.1 / 7.26 \\
MnasNet-Small & 0.88 / 974 & 246.9 / 65.1 & 580.6 / 6.82 \\
MambaOut-Femto & 6.27 / 2302 & 244.6 / 65.7 & 550.6 / 7.19 \\
MambaVision-T & 31.22 / 1604 & 302.9 / 53.1 & 650.0 / 6.09 \\
MobileOne-S1 & 3.67 / 2444 & 243.6 / 66.0 & 552.7 / 7.16 \\
PP-LCNet-1.0 & 1.80 / 1292 & 231.6 / 69.4 & 548.4 / 7.22 \\
RepGhostNet-0.5 & 1.16 / 814 & 247.2 / 65.0 & 562.5 / 7.04 \\
MobileNetV4-Conv-S & 2.62 / 988 & 244.8 / 65.7 & 567.7 / 6.97 \\
HardCoRe-NAS-A & 4.11 / 1850 & 243.2 / 66.1 & 582.7 / 6.79 \\
EfficientFormer-L1 & 11.48 / 2332 & 233.7 / 68.8 & 553.8 / 7.15 \\
EfficientViT-M0 & 2.18 / 658 & 246.3 / 65.2 & 583.9 / 6.78 \\
EfficientFormerV2-S1 & 5.78 / 2786 & 244.9 / 65.6 & 556.2 / 7.12 \\
InceptionNeXt-Tiny & 25.98 / 2446 & 245.6 / 65.4 & 555.7 / 7.12 \\
ConvFormer-S18 & 24.93 / 4284 & 231.2 / 69.5 & 508.7 / 7.78 \\
EfficientViT-B1 & 7.66 / 2616 & 246.5 / 65.2 & 569.4 / 6.95 \\
VOLO-D1 & 25.94 / 2596 & 238.6 / 67.4 & 455.0 / 8.70 \\
FocalNet-Small & 49.20 / 2296 & 193.7 / 83.0 & 463.9 / 8.53 \\
TinyViT-11M & 10.59 / 3732 & 244.4 / 65.8 & 561.1 / 7.05 \\
TinyViT-21M & 20.68 / 3268 & 231.7 / 69.4 & 544.5 / 7.27 \\
PVTv2-B0 & 3.44 / 1844 & 245.9 / 65.4 & 549.2 / 7.21 \\
ESE-VoVNet-19b-DW & 5.62 / 2970 & 245.1 / 65.6 & 562.7 / 7.03 \\
Swin-Small & 48.91 / 2398 & 159.2 / 100.9 & 656.9 / 6.03 \\
Twins-PCPVT-S & 23.64 / 2740 & 244.6 / 65.7 & 545.6 / 7.25 \\
Twins-SVT-S & 23.60 / 2268 & 232.6 / 69.1 & 537.9 / 7.36 \\
Visformer-Small & 39.53 / 1232 & 312.9 / 51.4 & 656.1 / 6.03 \\
MViTv2-Tiny & 23.48 / 4540 & 207.7 / 77.4 & 441.6 / 8.96 \\
MViTv2-Small & 34.18 / 5426 & 140.2 / 114.6 & 331.4 / 11.94 \\
GCViT-Tiny & 27.76 / 4568 & 127.1 / 126.4 & 374.3 / 10.57 \\
CoAtNet-1 & 41.03 / 3362 & 306.7 / 52.4 & 651.2 / 6.08 \\
Hiera-Tiny & 27.22 / 2940 & 242.0 / 66.4 & 556.0 / 7.12 \\
Hiera-Small & 34.32 / 3272 & 245.3 / 65.5 & 494.2 / 8.01 \\
DeiT-Small/16 & 21.70 / 1420 & 243.5 / 66.0 & 575.1 / 6.88 \\
ViT-Small/16 & 21.70 / 1420 & 245.5 / 65.5 & 575.2 / 6.88 \\
ConViT-Tiny & 5.54 / 2506 & 238.2 / 67.5 & 543.3 / 7.28 \\
CAFormer-S18 & 24.50 / 3924 & 247.0 / 65.1 & 522.5 / 7.58 \\
ViT-Tiny/16 & 5.54 / 1150 & 243.7 / 65.9 & 574.1 / 6.89 \\
ResNet-50 & 23.71 / 2090 & 244.8 / 65.7 & 561.7 / 7.05 \\
ResNet-101 & 42.71 / 1676 & 298.2 / 53.9 & 639.7 / 6.19 \\
DenseNet-121 & 7.06 / 3930 & 243.6 / 66.0 & 557.9 / 7.09 \\
EfficientNetV2-S & 20.31 / 2656 & 242.0 / 66.4 & 543.8 / 7.28 \\
RegNetY-8GF (timm) & 37.57 / 2262 & 303.4 / 53.0 & 633.1 / 6.25 \\
MobileNetV3-Large & 4.33 / 1774 & 241.5 / 66.6 & 551.5 / 7.18 \\
ConvNeXt-Tiny & 27.90 / 2412 & 241.9 / 66.4 & 560.7 / 7.06 \\
ConvNeXtV2-Tiny & 27.94 / 3780 & 219.2 / 73.3 & 502.2 / 7.88 \\
ViT-B/16 & 85.88 / 1986 & 290.0 / 55.4 & 575.6 / 6.88 \\
DeiT III-Small/16 & 21.71 / 1518 & 245.1 / 65.6 & 563.2 / 7.03 \\
Swin-Tiny & 27.60 / 2928 & 241.8 / 66.5 & 556.9 / 7.11 \\
RegNetY-3.2GF & 18.07 / 2112 & 240.8 / 66.7 & 541.5 / 7.31 \\
PVTv2-B2 & 24.90 / 3228 & 233.6 / 68.8 & 523.3 / 7.56 \\
MaxViT-Tiny & 30.46 / 3296 & 78.4 / 205.1 & 335.1 / 11.81 \\
EVA-02-Tiny & 5.52 / 1700 & 245.3 / 65.5 & 580.7 / 6.82 \\
EfficientFormerV2-S2 & 12.19 / 4028 & 243.8 / 65.9 & 450.4 / 8.79 \\
MAE ViT-B/16 & 85.88 / 1944 & 222.7 / 72.2 & 267.9 / 14.77 \\
DPN-68b & 11.86 / 3020 & 244.3 / 65.8 & 534.9 / 7.40 \\
GeRNet-S & 6.44 / 1060 & 232.7 / 69.1 & 544.8 / 7.27 \\
RepVGG-A1 & 12.94 / 1654 & 242.7 / 66.2 & 551.8 / 7.17 \\
SigLIP2 Base patch16-224 & 92.96 / 2064 & 183.1 / 87.8 & 310.9 / 12.73 \\
MambaOut-Tiny & 24.47 / 3458 & 243.4 / 66.0 & 539.5 / 7.34 \\
RepViT-M1.0 & 6.49 / 2728 & 234.6 / 68.5 & 548.4 / 7.22 \\
\end{longtable}
\par\smallskip
{\raggedright\footnotesize\textit{Model references.}
DaViT-Tiny~\citep{ding2022davit}; DLA-34~\citep{yu2018deep}; HRNet-W18~\citep{wang2021hrnet};
ConvMixer-768/32~\citep{trockman2023convmixer}; ResNeSt-26D~\citep{zhang2022resnest};
FBNetV3-B~\citep{dai2021fbnetv3}; ConvNeXt-Small/Tiny~\citep{liu2022convnet};
Sequencer2D-L~\citep{tatsunami2022sequencer}; XCiT-S12/8~\citep{ali2021xcit};
Res2Net-50~\citep{gao2021res2net}; EfficientNet-B0~\citep{tan2019efficientnet};
PoolFormer-S12~\citep{yu2022metaformer}; GCViT-XTiny/Tiny~\citep{hatamizadeh2023gcvit};
PiT-Tiny/Small~\citep{heo2021pit}; RegNetY-1.6GF/3.2GF/8GF~\citep{radosavovic2020designing};
gMLP-S16~\citep{liu2021gmlp}; ReXNet-1.0~\citep{han2020rexnet};
MLP-Mixer-B/16~\citep{tolstikhin2021mlpmixer}; GhostNetV2-1.0~\citep{tang2022ghostnetv2};
RepGhostNet-0.5/1.0~\citep{chen2022repghost}; MnasNet-Small~\citep{tan2019mnasnet};
MambaOut-Femto/Tiny~\citep{yu2025mambaout}; MambaVision-T~\citep{hatamizadeh2025mambavision};
MobileOne-S1~\citep{vasu2023mobileone}; PP-LCNet-1.0~\citep{cui2021pplcnet};
MobileNetV4-Conv-S~\citep{qin2024mobilenetv4}; HardCoRe-NAS-A~\citep{nayman2021hardcore};
EfficientFormer-L1~\citep{li2022efficientformer}; EfficientViT-M0~\citep{cai2023efficientvit};
EfficientFormerV2-S1/S2~\citep{li2023efficientformerv2}; InceptionNeXt-Tiny~\citep{yu2024inceptionnext};
ConvFormer-S18/CAFormer-S18~\citep{yu2024metaformerbaselines}; EfficientViT-B1~\citep{liu2023efficientvit};
VOLO-D1~\citep{yuan2023volo}; FocalNet-Small~\citep{yang2022focalnet};
TinyViT-11M/21M~\citep{wu2022tinyvit}; PVTv2-B0/B2~\citep{wang2022pvtv2};
ESE-VoVNet-19b-DW~\citep{lee2019esevovnet}; Swin-Small/Tiny~\citep{liu2021swin};
Twins-PCPVT-S/SVT-S~\citep{chu2021twins}; Visformer-Small~\citep{chen2021visformer};
MViTv2-Tiny/Small~\citep{li2022mvitv2}; CoAtNet-1~\citep{dai2021coatnet};
Hiera-Tiny/Small~\citep{ryali2023hiera}; DeiT-Small/16~\citep{touvron2021deit};
ViT-Tiny/16, ViT-Small/16, and ViT-B/16~\citep{dosovitskiy2021image};
ConViT-Tiny~\citep{dascoli2021convit}; ResNet-50/101~\citep{he2016deep};
DenseNet-121~\citep{huang2017densely}; EfficientNetV2-S~\citep{tan2021efficientnetv2};
MobileNetV3-Large~\citep{howard2019searching}; ConvNeXtV2-Tiny~\citep{woo2023convnextv2};
DeiT III-Small/16~\citep{touvron2022deit3}; MaxViT-Tiny~\citep{tu2022maxvit};
EVA-02-Tiny~\citep{fang2024eva02}; MAE ViT-B/16~\citep{he2022mae};
DPN-68b~\citep{chen2017dpn}; GeRNet-S~\citep{lin2020gernet}; RepVGG-A1~\citep{ding2021repvgg};
SigLIP2 Base patch16-224~\citep{tschannen2025siglip2}; RepViT-M1.0~\citep{wang2024repvit}.\par}
\end{small}

\subsection{Per-Class Behavior Across the Breadth Scan}
\label{app:breadth-class-analysis}

\begin{figure}[htbp]
  \centering
  \hcenteredincludegraphics[width=\linewidth]{-8.0pt}{8.0pt}{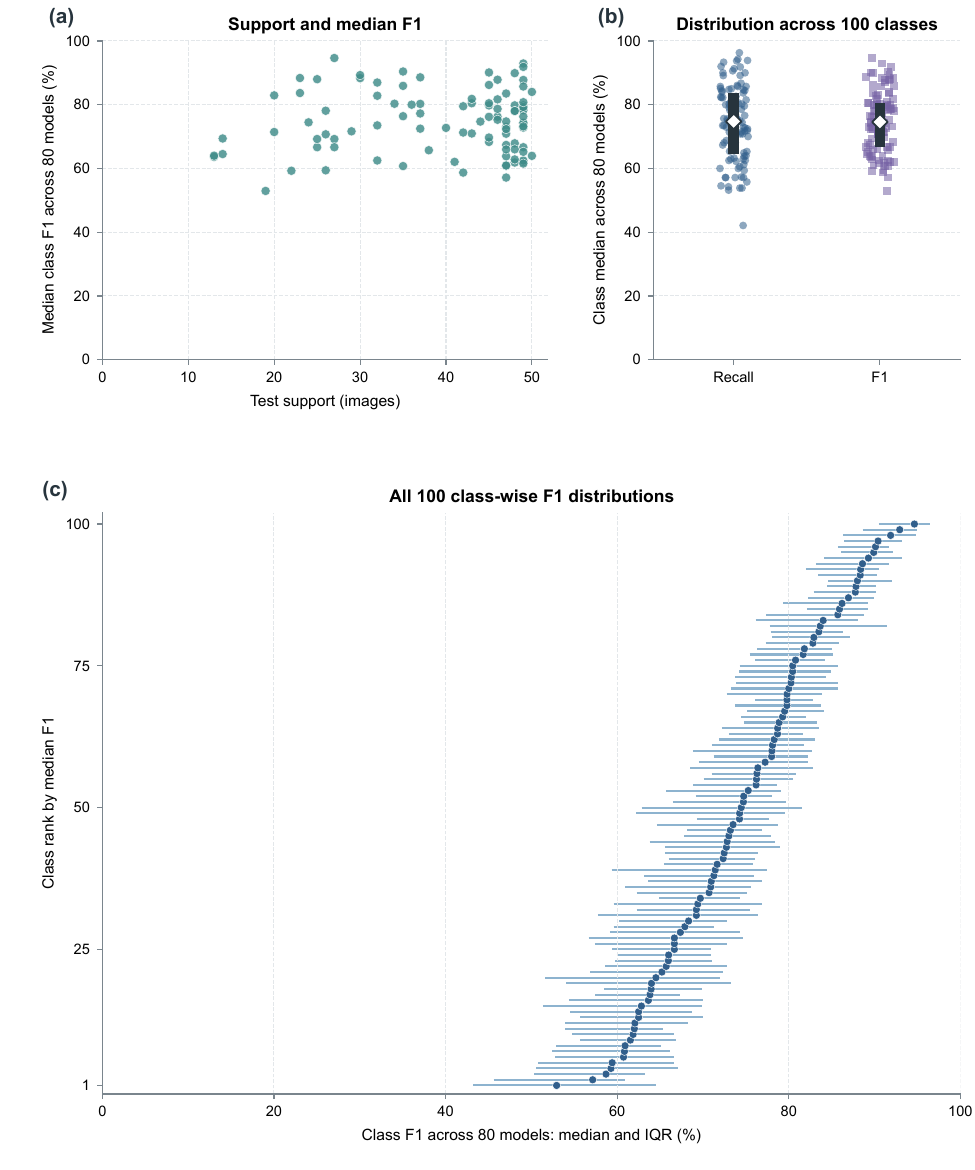}
  \caption{Per-class behavior across the 80-backbone scan. Panel (a) relates test support to median class F1. Panel (b) shows the distributions of all 100 class-level median Recall and F1 values; thick intervals denote IQRs and hollow diamonds denote medians. Panel (c) shows median F1 and IQR for every class, ranked by median F1.}
  \label{fig:breadth-classes}
\end{figure}

Figure~\ref{fig:breadth-classes} aggregates the 8,000 model--class records into a class-centered view. It relates test support to median class F1, summarizes the distribution of class-level median Recall and F1, and retains the median and interquartile F1 range for all 100 classes.

\subsection{Calibration and Inference Profiling}
\label{app:calibration-efficiency}

Two five-model analyses complement accuracy-based comparisons. Calibration uses held-out classification predictions and 15 equal-width confidence bins. Expected calibration error (ECE) and maximum calibration error (MCE) follow the standard bin-based definitions~\citep{guo2017calibration}. Selective prediction uses the risk--coverage formulation~\citep{geifman2017selective}, and the area under this curve (AURC) provides a scalar summary~\citep{ding2020uncertainty}. Tables~\ref{tab:calibration} and~\ref{tab:selective-prediction} report these two views separately.

Inference profiling uses the seed-41, 100-class checkpoint for each model; RepViT-M1.0 uses timm's fused inference graph. Measurements were collected on one NVIDIA GeForce \mbox{RTX 4080 SUPER} in FP32 with $224\times224$ RGB tensors. Each timing block used five warm-up iterations followed by 20 synchronized iterations. Table~\ref{tab:inference-complexity} reports parameter count, FLOPs, and peak CUDA memory; Table~\ref{tab:inference-efficiency} reports batch-one and batch-32 timing. MobileNetV3-Large provides the highest batch-32 throughput in this profile, while DINOv2 ViT-B/14 has the largest operation count and memory footprint.

\begin{table}[htbp]
\centering
\caption{Calibration measures for five seed-41 checkpoints on held-out data. Top-1, confidence, ECE, and MCE are percentages.}
\label{tab:calibration}
\begin{small}
\begin{tabular}{@{}>{\raggedright\arraybackslash}p{0.35\textwidth} r r r r r@{}}
\toprule
Model & Top-1 & Conf. & NLL & ECE & MCE \\
\midrule
ResNet-50 & 79.76 & 89.42 & 0.910 & 9.65 & 29.06 \\
ConvNeXtV2-Tiny & 77.97 & 95.99 & 1.736 & 18.07 & 51.58 \\
SwinV2-Tiny & 56.37 & 29.23 & 2.025 & 27.13 & 46.50 \\
MaxViT-Tiny & 84.21 & 95.27 & 0.878 & 11.06 & 39.28 \\
DINOv2 ViT-B/14 & 83.73 & 95.42 & 1.085 & 11.69 & 38.88 \\
\bottomrule
\end{tabular}
\par\smallskip
\parbox{\linewidth}{\raggedright\footnotesize\textit{Model references.}
ResNet-50~\citep{he2016deep}; ConvNeXtV2-Tiny~\citep{woo2023convnextv2};
SwinV2-Tiny~\citep{liu2022swinv2}; MaxViT-Tiny~\citep{tu2022maxvit};
DINOv2 ViT-B/14~\citep{oquab2024dinov2}.}
\end{small}
\end{table}

\begin{table}[htbp]
\centering
\caption{Selective-prediction measures for the same five seed-41 checkpoints. Risk and accuracy at 95\% coverage are percentages. AURC is reported as a raw, unitless area.}
\label{tab:selective-prediction}
\begin{small}
\begin{tabular}{@{}>{\raggedright\arraybackslash}p{0.45\textwidth} r r r@{}}
\toprule
Model & Risk@95 & Acc.@95 & AURC \\
\midrule
ResNet-50 & 17.28 & 82.72 & 0.047 \\
ConvNeXtV2-Tiny & 19.17 & 80.83 & 0.058 \\
SwinV2-Tiny & 41.53 & 58.47 & 0.205 \\
MaxViT-Tiny & 12.79 & 87.21 & 0.033 \\
DINOv2 ViT-B/14 & 13.24 & 86.76 & 0.033 \\
\bottomrule
\end{tabular}
\par\smallskip
\parbox{\linewidth}{\raggedright\footnotesize\textit{Model references.}
ResNet-50~\citep{he2016deep}; ConvNeXtV2-Tiny~\citep{woo2023convnextv2};
SwinV2-Tiny~\citep{liu2022swinv2}; MaxViT-Tiny~\citep{tu2022maxvit};
DINOv2 ViT-B/14~\citep{oquab2024dinov2}.}
\end{small}
\end{table}

\begin{table}[htbp]
\centering
\caption{Model complexity and peak CUDA memory for FP32 inference on an RTX~4080~SUPER with $224\times224$ RGB inputs. RepViT-M1.0 uses timm's fused inference graph in this profile, whereas the 80-backbone scan retains its two-head distillation graph.}
\label{tab:inference-complexity}
\begin{small}
\begin{tabular}{@{}>{\raggedright\arraybackslash}p{0.45\textwidth} r r r@{}}
\toprule
Model & Params (M) & FLOPs (G) & Peak MB \\
\midrule
ResNet-50 & 23.71 & 8.17 & 453.44 \\
MobileNetV3-Large & 4.33 & 0.43 & 263.69 \\
EfficientFormerV2-S2 & 12.19 & 2.50 & 298.19 \\
DINOv2 ViT-B/14 & 85.80 & 43.90 & 642.96 \\
RepViT-M1.0 & 6.41 & 2.21 & 246.75 \\
\bottomrule
\end{tabular}
\par\smallskip
\parbox{\linewidth}{\raggedright\footnotesize\textit{Model references.}
ResNet-50~\citep{he2016deep}; MobileNetV3-Large~\citep{howard2019searching};
EfficientFormerV2-S2~\citep{li2023efficientformerv2}; DINOv2 ViT-B/14~\citep{oquab2024dinov2};
RepViT-M1.0~\citep{wang2024repvit}.}
\end{small}
\end{table}

\begin{table}[htbp]
\centering
\caption{Synchronized FP32 inference timing on an RTX~4080~SUPER. Values are medians across three repetitions, each averaged over 20 synchronized timed iterations after five warm-up iterations.}
\label{tab:inference-efficiency}
\begin{small}
\begin{tabular}{@{}>{\raggedright\arraybackslash}p{0.36\textwidth} r r r r@{}}
\toprule
Model & B1 lat. (ms) & B1 img/s & B32 lat. (ms) & B32 img/s \\
\midrule
ResNet-50 & 6.03 & 165.8 & 17.29 & 1850.6 \\
MobileNetV3-Large & 6.31 & 158.4 & 6.59 & 4857.2 \\
EfficientFormerV2-S2 & 22.38 & 44.7 & 28.47 & 1123.8 \\
DINOv2 ViT-B/14 & 5.57 & 179.4 & 66.13 & 483.9 \\
RepViT-M1.0 & 7.05 & 141.8 & 7.06 & 4534.9 \\
\bottomrule
\end{tabular}
\par\smallskip
\parbox{\linewidth}{\raggedright\footnotesize\textit{Model references.}
ResNet-50~\citep{he2016deep}; MobileNetV3-Large~\citep{howard2019searching};
EfficientFormerV2-S2~\citep{li2023efficientformerv2}; DINOv2 ViT-B/14~\citep{oquab2024dinov2};
RepViT-M1.0~\citep{wang2024repvit}.}
\end{small}
\end{table}

\section{Complementary Recognition and Retrieval Analyses}
\label{app:recognition-retrieval}

\subsection{Complementary Retrieval Measures}
\label{app:retrieval-measures}

\begin{figure}[htbp]
  \centering
  \hcenteredincludegraphics[width=\linewidth]{-18.0pt}{18.0pt}{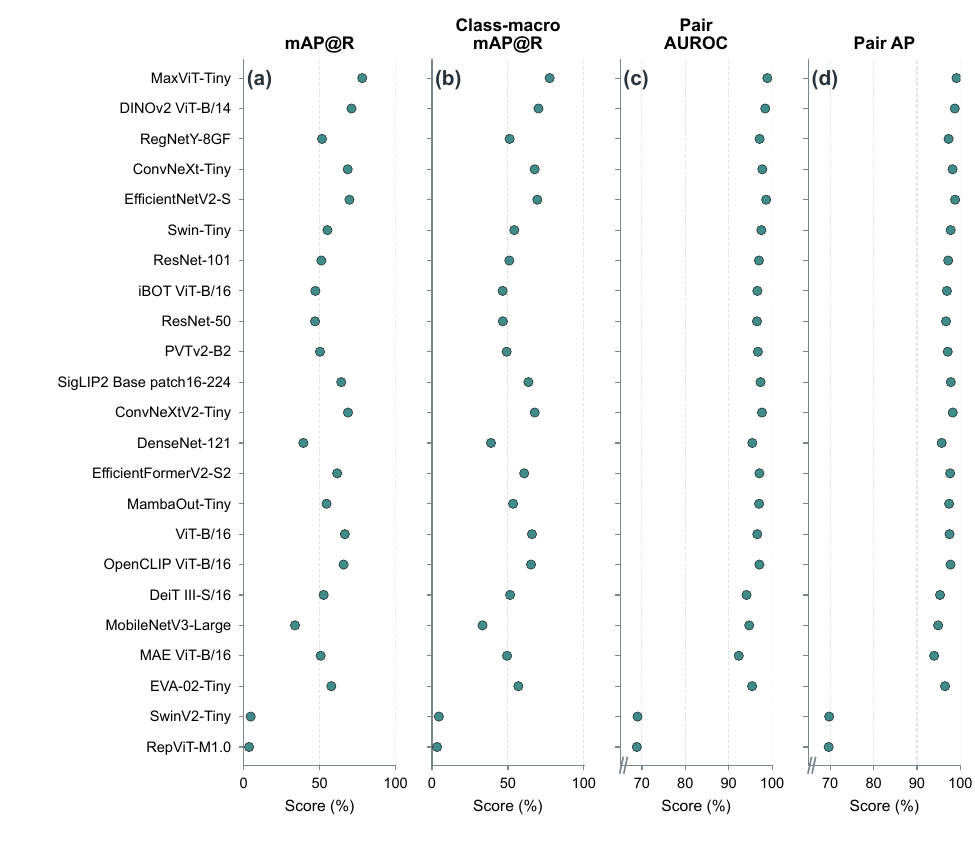}
  \caption{Complementary held-out retrieval results for 23 classifiers. Panels (a)--(d) report mAP@R, class-macro mAP@R, Pair AUROC, and Pair AP. The pairwise measures use 79,160 fixed query--gallery comparisons per classifier and axis limits of 65--100, with labeled ticks beginning at 70.}
  \label{fig:retrieval-diagnostics}
\end{figure}

The retrieval protocol uses the 3,958 test images as queries and the 32,143 training images as the gallery. For every classifier, the retrieval vector is the tensor supplied directly to the trained 100-class head, captured before the head operation and then L2-normalized. Table~\ref{tab:heldout-retrieval-23x3} reports three-seed R@1, R@5, and class-macro mAP@R. Panels (a)--(d) of Figure~\ref{fig:retrieval-diagnostics} complement those aggregate results with ordinary mAP@R~\citep{musgrave2020metric} and pairwise discrimination measures from a separate fixed held-out diagnostic. For each query, a deterministic image-ID-based procedure fixes 10 same-class positive and 10 different-class negative gallery images, yielding $3{,}958\times20=79{,}160$ pairs with equal positive and negative prevalence. This diagnostic is neither the full query--gallery Cartesian product nor a hard-negative sample. The pairwise panels use axis limits of 65--100, with labeled ticks beginning at 70.

\subsection{Few-Shot Prototype Readout}
\label{app:few-shot-readout}

We compare five fully trained seed-41 checkpoints using the same frozen support episodes. The set contains ResNet-50, ConvNeXtV2-Tiny, SwinV2-Tiny, DINOv2 ViT-B/14, and SigLIP2 Base patch16-224. Each representation is learned during its full 100-epoch supervised run; the few-shot stage constructs class prototypes without updating the encoder.

\begin{figure}[htbp]
  \centering
  \hcenteredincludegraphics[width=\linewidth]{-5.2pt}{5.2pt}{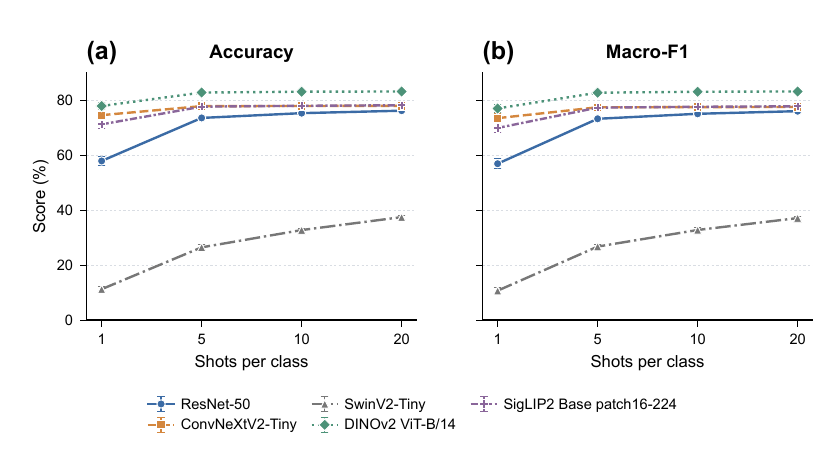}
  \caption{Few-shot prototype readout for five fully trained fixed checkpoints. Panels (a) and (b) report Accuracy and Macro-F1 across 1, 5, 10, and 20 support images per class. Points show means over 90 frozen episodes, and error bars denote episode-level sample SD. All checkpoints use seed 41, and every episode queries the same 3,958-image held-out test set.}
  \label{fig:few-shot-prototype-readout}
\end{figure}

For each checkpoint, we evaluate 1, 5, 10, and 20 support images per class over 90 fixed episodes. Support images come from distinct training groups, and every episode queries the fixed 3,958-image held-out test set. We average the L2-normalized support embeddings within each class, normalize the resulting prototype, and assign queries by cosine similarity. Figure~\ref{fig:few-shot-prototype-readout} reports the episode means and sample SDs for Accuracy and Macro-F1.

DINOv2 records the highest mean at every shot count for both metrics. At 20 shots, it reaches 83.35\% Accuracy and 83.36\% Macro-F1. SwinV2-Tiny shows the largest increase, gaining 26.25 and 26.44 percentage points from one to 20 shots.

\subsection{Taxonomy-Aware Recognition}
\label{app:taxonomy-recognition}

\begin{figure}[htbp]
  \centering
  \hcenteredincludegraphics[width=\linewidth]{-15.4pt}{15.4pt}{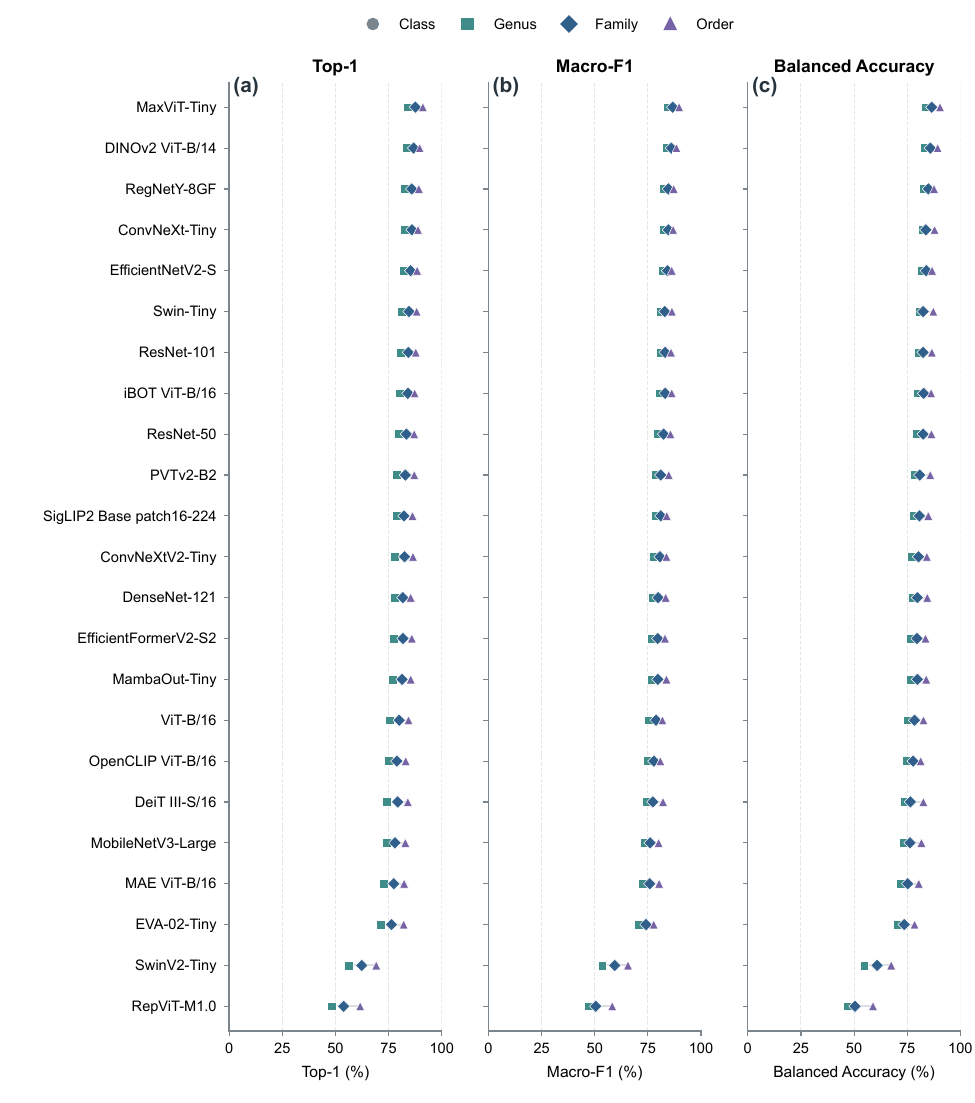}
  \caption{Taxonomy-aware recognition for 23 models. Panels report Top-1, Macro-F1, and Balanced Accuracy after projecting the same class predictions to class, genus, family, and order. Connecting segments trace the four taxonomic levels for each model.}
  \label{fig:taxonomy-projection}
\end{figure}

Top-1 class predictions are projected to genus, family, and order. This separates exact-class errors from errors within a nearby taxonomic group. Figure~\ref{fig:taxonomy-projection} aligns Top-1, Macro-F1, and Balanced Accuracy across class, genus, family, and order for all 23 models, making the retained higher-rank information directly comparable across metrics.

\subsection{Family-Auxiliary Supervision}
\label{app:family-auxiliary}

Taxonomic projection measures information retained after species prediction. A separate paired comparison tests whether family supervision changes recognition performance. We add a 55-way family head to the ResNet-50 benchmark configuration while retaining the 100-way species head as the primary output. The species and family losses receive equal weight, and both heads use the same augmented image. Each paired run follows the benchmark training schedule and selects its checkpoint by validation species Macro-F1.

Table~\ref{tab:family-auxiliary} compares the three-run held-out results. The family head reaches 82.11$\pm$0.46 accuracy, 81.10$\pm$0.71 Macro-F1, and 80.31$\pm$0.78 Balanced Accuracy on its 55-way output. Under the tested objective, family supervision does not improve species classification: Top-1, Macro-F1, and Balanced Accuracy decrease by 1.34, 1.22, and 1.20 percentage points. Retrieval changes direction across metrics. R@1 and R@5 decrease by 1.02 and 1.35 points, whereas class-macro mAP@R increases by 1.16 points.

\begin{table}[H]
\centering
\scriptsize
\setlength{\tabcolsep}{2pt}
\caption{Three-run held-out results with and without family-auxiliary supervision. Values are mean $\pm$ sample SD in percentage points; paired changes are auxiliary minus baseline. Retrieval mAP@R is class-macro mAP@R. Paired changes are computed from the unrounded seed-level results before display rounding.}
\label{tab:family-auxiliary}
\begin{tabular}{p{1.40in}rrrrrrr}
\toprule
Setting & Top-1 & Top-5 & Macro-F1 & Bal.\ acc. & R@1 & R@5 & mAP@R \\
\midrule
ResNet-50 & 79.82$\pm$0.05 & 93.92$\pm$0.08 & 79.61$\pm$0.12 & 79.47$\pm$0.14 & 77.75$\pm$0.23 & 87.70$\pm$0.42 & 46.61$\pm$0.93 \\
$+$ family auxiliary head & 78.48$\pm$0.55 & 93.41$\pm$0.23 & 78.39$\pm$0.68 & 78.27$\pm$0.70 & 76.73$\pm$0.57 & 86.36$\pm$0.50 & 47.77$\pm$0.16 \\
Paired change & $-1.34$ & $-0.51$ & $-1.22$ & $-1.20$ & $-1.02$ & $-1.35$ & $+1.16$ \\
\bottomrule
\end{tabular}
\par\smallskip
\parbox{\linewidth}{\raggedright\footnotesize\textit{Model reference.}
ResNet-50~\citep{he2016deep}.}
\end{table}

\subsection{Scene-Stratified Recognition}
\label{app:scene-recognition}

\begin{figure}[htbp]
  \centering
  \hcenteredincludegraphics[width=\linewidth]{-15.4pt}{15.4pt}{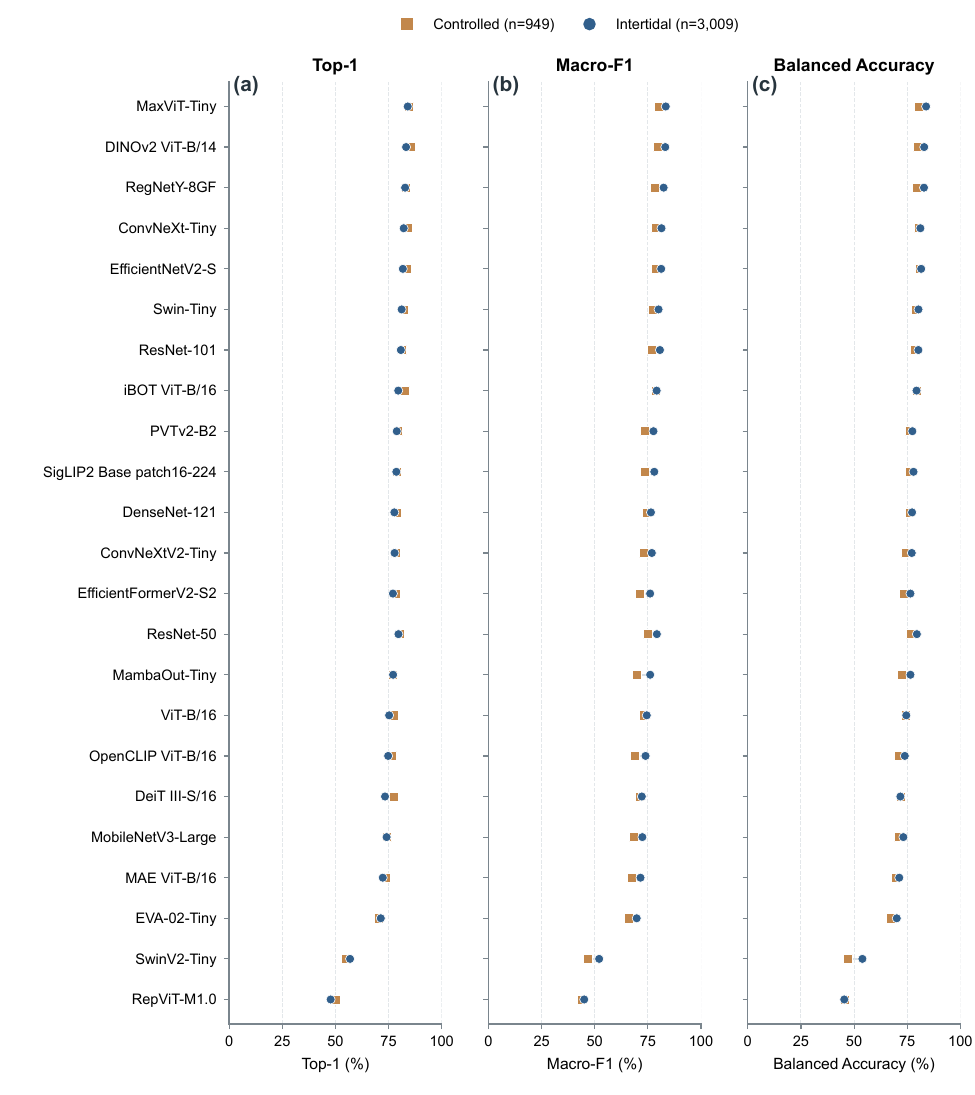}
  \caption{Recognition results stratified by scene type for the fixed seed-41 checkpoint of each of the 23 models. Panels report Top-1, Macro-F1, and Balanced Accuracy, with paired points connecting the controlled ($n=949$) and intertidal ($n=3{,}009$) results for each model.}
  \label{fig:scene-strata}
\end{figure}

The held-out partition contains 949 controlled-scene images and 3,009 intertidal-scene images. Figure~\ref{fig:scene-strata} compares the two strata for the fixed seed-41 checkpoint of each model using Top-1, Macro-F1, and Balanced Accuracy. Paired points make the direction and magnitude of each within-model difference visible across the three metrics. The strata differ in subject presentation, background, class composition, and image source.

We also compare the 96 taxa represented in both strata. Controlled support ranges from 1 to 31 images per taxon, while intertidal support ranges from 4 to 48. For each model--seed run, recall is computed for every taxon and scene and macro-averaged across taxa. Table~\ref{tab:scene-matched-recall} reports the mean and sample SD over seeds 41, 42, and 43. This comparison fixes the taxon set and class weighting while retaining the observed images in each stratum.

\begin{table}[htbp]
\centering
\scriptsize
\setlength{\tabcolsep}{4.5pt}
\renewcommand{\arraystretch}{0.95}
\caption{Taxon-matched scene analysis over the 96 taxa represented in both strata, comprising 949 controlled images and 2,874 intertidal images. Recall is computed for each taxon and scene, macro-averaged across taxa within each seed, and reported as mean $\pm$ sample SD over seeds 41, 42, and 43. $\Delta$ is controlled minus intertidal within each seed and is summarized across seeds in percentage points.}
\label{tab:scene-matched-recall}
\begin{tabular}{@{}lccc@{}}
\toprule
Model & Controlled & Intertidal & $\Delta$ \\
\midrule
ResNet-50 & $77.64 \pm 1.28$ & $78.84 \pm 0.68$ & $-1.19 \pm 1.63$ \\
ResNet-101 & $78.59 \pm 0.44$ & $79.85 \pm 0.37$ & $-1.26 \pm 0.43$ \\
DenseNet-121 & $74.44 \pm 1.85$ & $76.80 \pm 0.40$ & $-2.36 \pm 1.47$ \\
EfficientNetV2-S & $80.68 \pm 2.15$ & $81.86 \pm 0.63$ & $-1.17 \pm 2.73$ \\
RegNetY-8GF & $80.05 \pm 0.56$ & $83.18 \pm 0.41$ & $-3.13 \pm 0.84$ \\
MobileNetV3-Large & $72.75 \pm 2.15$ & $73.61 \pm 0.57$ & $-0.86 \pm 1.59$ \\
ConvNeXt-Tiny & $80.36 \pm 0.11$ & $80.70 \pm 0.36$ & $-0.34 \pm 0.29$ \\
ConvNeXtV2-Tiny & $76.38 \pm 1.97$ & $77.02 \pm 0.47$ & $-0.65 \pm 2.21$ \\
ViT-B/16 & $74.09 \pm 1.26$ & $74.37 \pm 0.89$ & $-0.28 \pm 0.37$ \\
DeiT III-Small/16 & $72.79 \pm 2.24$ & $73.03 \pm 1.67$ & $-0.24 \pm 1.40$ \\
Swin-Tiny & $79.64 \pm 1.38$ & $80.93 \pm 0.67$ & $-1.29 \pm 0.82$ \\
SwinV2-Tiny & $39.58 \pm 7.47$ & $46.05 \pm 6.79$ & $-6.47 \pm 1.62$ \\
PVTv2-B2 & $76.52 \pm 0.87$ & $78.25 \pm 1.38$ & $-1.73 \pm 0.51$ \\
MaxViT-Tiny & $81.22 \pm 0.71$ & $84.50 \pm 0.45$ & $-3.28 \pm 0.27$ \\
EVA-02-Tiny & $67.42 \pm 1.21$ & $69.89 \pm 0.30$ & $-2.47 \pm 1.13$ \\
EfficientFormerV2-S2 & $75.57 \pm 1.92$ & $76.55 \pm 0.39$ & $-0.98 \pm 2.02$ \\
MAE ViT-B/16 & $69.32 \pm 0.48$ & $71.49 \pm 0.15$ & $-2.17 \pm 0.61$ \\
iBOT ViT-B/16 & $78.44 \pm 1.17$ & $78.42 \pm 1.56$ & $0.02 \pm 0.59$ \\
DINOv2 ViT-B/14 & $79.51 \pm 0.51$ & $81.93 \pm 0.98$ & $-2.42 \pm 0.66$ \\
OpenCLIP ViT-B/16 & $71.77 \pm 2.11$ & $73.46 \pm 0.45$ & $-1.70 \pm 2.56$ \\
SigLIP2 Base patch16-224 & $75.39 \pm 1.84$ & $77.24 \pm 0.54$ & $-1.85 \pm 1.68$ \\
MambaOut-Tiny & $74.42 \pm 2.03$ & $76.18 \pm 0.55$ & $-1.76 \pm 2.58$ \\
RepViT-M1.0 & $47.73 \pm 2.04$ & $49.11 \pm 3.55$ & $-1.38 \pm 2.08$ \\
\bottomrule
\end{tabular}
\end{table}

Across the three-seed means, controlled-scene recall has a lower point estimate for 22 of the 23 models; iBOT ViT-B/16 differs by 0.02 percentage points in the opposite direction.

\subsection{Shared-Taxon Transfer to BackHome19K}
\label{app:backhome-transfer}

We evaluate the three ResNet-50 benchmark checkpoints on taxonomically matched BackHome19K images without further training. Scientific binomials are parsed mechanically from filenames and matched by case-normalized exact equality. This procedure yields 227 images from eight shared taxa. Cross-dataset perceptual-hash screening removes three images from the shared set, leaving 224 images from eight taxa. None of the removed images belongs to the official test subset, which contains 25 images from seven shared taxa.

Table~\ref{tab:backhome-transfer} reports the official-test results in two prediction spaces. The original output retains all 100 ShellfishNet classes. The restricted output masks nonshared classes before ranking the seven shared taxa. On this subset, restriction raises mean Top-1 accuracy from 68.00\% to 96.00\% and Top-5 accuracy from 88.00\% to 100.00\%. Because the official subset contains only 25 images from seven shared taxa, this experiment is treated as a limited shared-taxon diagnostic. The restricted result measures discrimination after nonshared classes are masked and is not directly comparable with the original 100-class result.

\begin{table}[H]
\centering
\small
\setlength{\tabcolsep}{5pt}
\caption{Frozen ResNet-50 transfer to the 25-image BackHome19K official-test subset. Values are mean $\pm$ sample SD over three checkpoints, in percentage points.}
\label{tab:backhome-transfer}
\begin{tabular}{p{2.35in}rrrr}
\toprule
Prediction space & Top-1 & Top-5 & Macro-F1 & Bal.\ acc. \\
\midrule
Original 100-class output & 68.00$\pm$4.00 & 88.00$\pm$0.00 & 71.77$\pm$3.78 & 63.89$\pm$5.50 \\
Restricted to seven shared taxa & 96.00$\pm$0.00 & 100.00$\pm$0.00 & 86.94$\pm$7.07 & 88.89$\pm$5.50 \\
\bottomrule
\end{tabular}
\par\smallskip
\parbox{\linewidth}{\raggedright\footnotesize\textit{Dataset and model references.}
BackHome19K~\citep{valverde2025backhome}; ResNet-50~\citep{he2016deep}.}
\end{table}

\subsection{Frozen Zero-Shot Classification and Sensitivity}
\label{app:zero-shot-classification}

Table~\ref{tab:zero-shot} records frozen zero-shot classification with four scientific-name templates. OpenCLIP ViT-B/16 and SigLIP2 Base patch16-224 use their released pretrained representations. The OpenCLIP evaluation uses \texttt{open-clip-torch==2.24.0}. We also report BioCLIP 2 as an exploratory zero-shot baseline. Its Top-1 accuracy is 87.97\%, above the highest 23-model fine-tuned mean of 84.39\%, but this is not a leakage-controlled or like-for-like comparison. BioCLIP 2 was trained on TreeOfLife-200M, a 214-million-image biological dataset, with additional LAION-2B experience replay~\citep{gu2025bioclip2}. The ShellfishNet test partition is drawn mainly from iNaturalist and also includes GBIF, Wikimedia Commons, iDigBio, and the Encyclopedia of Life. TreeOfLife-200M documentation lists GBIF and the Encyclopedia of Life among its core providers, so provider-level source overlap is confirmed for part of the evaluation data. The released BioCLIP 2 checkpoint does not provide a complete training-image manifest or source identifiers, so exact image-level and species-level overlap cannot be measured here; we neither assert nor exclude such overlap. The result is therefore kept outside the fine-tuned ranking and is interpreted only as an exploratory, overlap-sensitive diagnostic rather than evidence of strong external generalization. All three evaluations use the same fixed test partition without training or test-time augmentation. Each row comes from one deterministic frozen evaluation, so no seed dispersion is estimated.

\begin{small}
\begin{longtable}{>{\raggedright\arraybackslash}p{0.40\textwidth} r r r r}
\caption{Frozen zero-shot classification on the 3,958-image test partition. All values are percentages from one evaluation with four scientific-name templates. BioCLIP 2 is an exploratory baseline.}\label{tab:zero-shot}\\
\toprule
Encoder & Top-1 & Top-5 & Macro-F1 & Bal.\ acc. \\
\midrule
\endfirsthead
\multicolumn{5}{l}{\tablename~\thetable\ (continued)}\\
\toprule
Encoder & Top-1 & Top-5 & Macro-F1 & Bal.\ acc. \\
\midrule
\endhead
\midrule
\multicolumn{5}{r}{Continued on next page}\\
\endfoot
\bottomrule
\endlastfoot
OpenCLIP ViT-B/16 & 2.98 & 10.11 & 1.44 & 2.88 \\
SigLIP2 Base patch16-224 & 0.71 & 4.95 & 0.25 & 1.08 \\
BioCLIP 2 (ViT-L/14) & 87.97 & 97.98 & 87.65 & 87.54 \\
\end{longtable}
\par\smallskip
\parbox{\linewidth}{\raggedright\footnotesize\textit{Model references.}
OpenCLIP ViT-B/16~\citep{cherti2023openclip};
SigLIP2 Base patch16-224~\citep{tschannen2025siglip2};
BioCLIP 2~\citep{gu2025bioclip2}.}
\end{small}

\begin{figure}[htbp]
  \centering
  \hcenteredincludegraphics[width=\linewidth]{-3.5pt}{3.5pt}{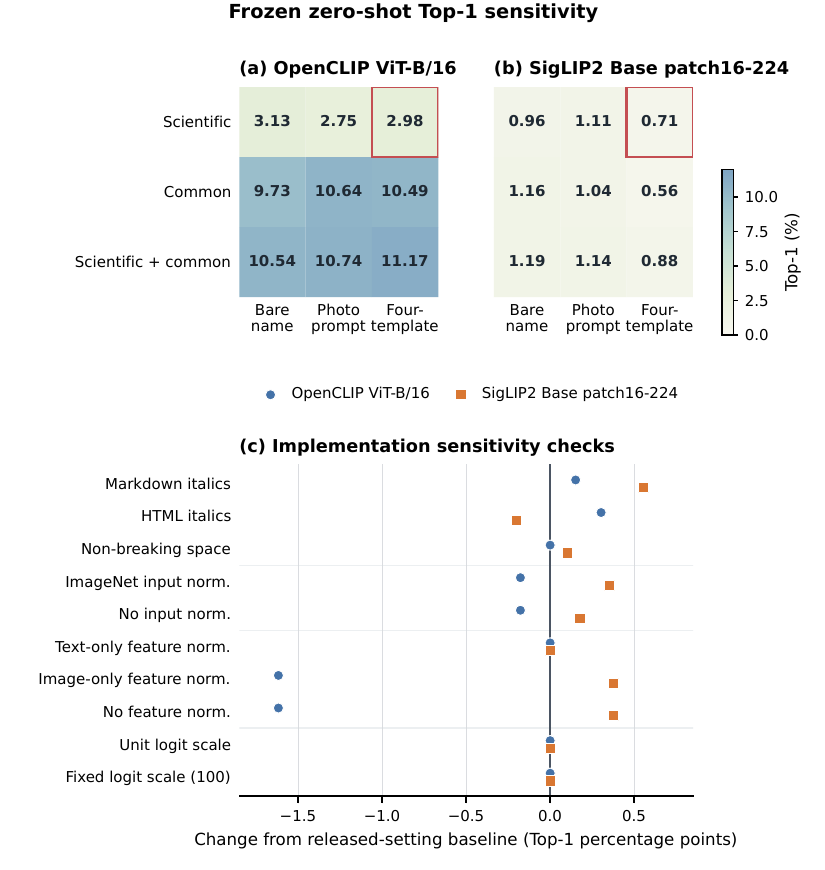}
  \caption{Frozen zero-shot sensitivity on the fixed 3,958-image test partition. Panels (a) and (b) show Top-1 for the nine name--prompt combinations. Both heatmaps use the same 0--12\% color scale, and exact values are printed in each cell. The outlined cells corresponding to scientific names and the four-template prompt mark the released-setting baselines. Panel (c) reports Top-1 changes under tokenizer, input-normalization, feature-normalization, and logit-scale checks. All non-baseline conditions are diagnostic probes rather than selected benchmark entries.}
  \label{fig:zero-shot-diagnostics}
\end{figure}

The sensitivity study keeps model weights and test images fixed. It evaluates 19 conditions per model. A $3\times3$ matrix crosses scientific, common, and combined class names with bare, photo, and four-template prompts. The bare condition uses \texttt{\{name\}}, and the photo condition uses \texttt{a photo of \{name\}.} The four-template condition averages \texttt{a photo of \{name\}.}, \texttt{a close-up photo of \{name\}.}, \texttt{a natural history image of \{name\}.}, and \texttt{a shellfish specimen of \{name\}.} Ten other checks vary tokenizer formatting, input normalization, feature normalization, or logit scale. Across the name--prompt matrix, OpenCLIP Top-1 ranges from 2.75\% to 11.17\%. SigLIP2 Base patch16-224 ranges from 0.56\% to 1.19\%. Figure~\ref{fig:zero-shot-diagnostics} shows the full matrix and the remaining checks. The released-setting values stay as the benchmark baselines; the other conditions serve as sensitivity probes.

\section{Complete Captioning Evaluation}
\label{app:caption-evaluation}

\subsection{Models and Common Captioning Protocol}
\label{app:caption-protocol}

The captioning benchmark uses a fixed subset of 512 test images covering all 100 classes and produces 5,120 model outputs. The ten open models span classic image-captioning systems and more recent vision--language models, as listed in Table~\ref{tab:caption-models}. Every model receives the same standardized $1024\times1024$ RGB image. Each source image is converted to RGB, resized with Lanczos interpolation while preserving its aspect ratio, centered on a white $1024\times1024$ canvas, and saved as a quality-95 JPEG before model-specific processing. Chat-based vision--language models receive a common English instruction. Florence-2 uses its native \texttt{\textless CAPTION\textgreater} task token, while the classic captioning models use their native deterministic image-to-text interfaces. The resulting scores compare complete inference pipelines, so architecture and task-interface effects remain combined.

\begin{small}
\begin{longtable}{>{\raggedright\arraybackslash}p{0.27\textwidth} >{\raggedright\arraybackslash}p{0.22\textwidth} l >{\raggedright\arraybackslash}p{0.31\textwidth}}
\caption{Open models in the 512-image captioning evaluation.}\label{tab:caption-models}\\
\toprule
Model & Family & Scale & Inference interface \\
\midrule
\endfirsthead
\multicolumn{4}{l}{\tablename~\thetable\ (continued)}\\
\toprule
Model & Family & Scale & Inference interface \\
\midrule
\endhead
\midrule
\multicolumn{4}{r}{Continued on next page}\\
\endfoot
\bottomrule
\endlastfoot
ViT-GPT2 & Classic captioning & $<1$B & Single-image deterministic captioning \\
BLIP-Large & Classic captioning & $<1$B & Single-image deterministic captioning \\
Florence-2 Base FT & Compact foundation model & $<1$B & Single-image prompted captioning \\
LLaVA-Med 1.5 7B & Biology-specialized vision--language model & 7B & Single-image chat inference \\
Qwen2.5-VL-3B & General vision--language model & 3B & Single-image chat inference \\
Qwen3-VL-4B & General vision--language model & 4B & Single-image chat inference \\
LLaVA-1.5-7B & General vision--language model & 7B & Single-image chat inference \\
BLIP-Base & Classic captioning & $<1$B & Single-image deterministic captioning \\
InternVL2.5-8B & General vision--language model & 8B & Single-image chat inference \\
Phi-3.5 Vision & Compact vision--language model & 4B & Single-image chat inference \\
\end{longtable}
\par\smallskip
\parbox{\linewidth}{\raggedright\footnotesize\textit{Model references.}
ViT-GPT2~\citep{nlpconnect2022vitgpt2,dosovitskiy2021image,radford2019gpt2};
BLIP-Large and BLIP-Base~\citep{li2022blip}; Florence-2 Base FT~\citep{xiao2024florence2};
LLaVA-Med 1.5 7B~\citep{li2023llavamed}; Qwen2.5-VL-3B~\citep{bai2025qwen25vl};
Qwen3-VL-4B~\citep{bai2025qwen3vl}; LLaVA-1.5-7B~\citep{liu2024llava15};
InternVL2.5-8B~\citep{chen2024internvl25}; Phi-3.5 Vision~\citep{abdin2024phi3}.}
\end{small}

Table~\ref{tab:caption-prompt} gives the common instruction used by chat-based vision--language models. Decoding uses batch size 1 and at most 48 new tokens. Prompted models use deterministic generation with \texttt{do\_sample=False} and temperature 0.0 where supported. Classic captioners use deterministic native generation with \texttt{no\_repeat\_ngram\_size=3}; model-specific end-of-sequence handling remains native. A human domain expert annotates and finalizes the reference captions, using GPT-5.6 Sol to assist with initial drafts. The expert inspects each image and revises every caption to follow the same one-sentence format.

\begin{small}
\begin{longtable}{>{\raggedright\arraybackslash}p{0.20\textwidth} >{\raggedright\arraybackslash}p{0.73\textwidth}}
\caption{Common instruction used by chat-based vision--language models, shown as its constituent requirements.}\label{tab:caption-prompt}\\
\toprule
Component & Instruction \\
\midrule
\endfirsthead
\multicolumn{2}{l}{\tablename~\thetable\ (continued)}\\
\toprule
Component & Instruction \\
\midrule
\endhead
\midrule
\multicolumn{2}{r}{Continued on next page}\\
\endfoot
\bottomrule
\endlastfoot
Role & You are a scientific image captioner for a marine biology image set. \\
Task & Write one English sentence describing the image. \\
Visible content & Mention the main subject, visible attributes, and marine scene context. \\
Taxonomic wording & If a confident common name, genus, or species name fits the image, include it once. \\
Style & Prefer concrete nouns and visible traits over guesses. \\
Conflict handling & Ignore any cue that conflicts with the image. \\
Length & Keep the caption within 30 words. \\
Output format & Return a single sentence only, with no bullets, line breaks, or explanation. \\
\end{longtable}
\end{small}

\subsection{Blinded Human Evaluation}
\label{app:caption-human-evaluation}

Two human reviewers evaluated the generated captions with model identities hidden. They scored subject correctness (M1), visual factuality (M2), grounded informativeness (M3), and language and format (M4), assigning 0, 1, or 2 for each criterion. The four ratings sum to an overall score from 0 to 8. A fixed seeded permutation assigns 256 primary images to each reviewer. Ten images are sampled from each half for reciprocal overlap scoring, yielding the 20-image, 200-output agreement set. Model identities appear as anonymous slots, and outputs are shown in randomized order. Every model output received one primary score, giving 512 scores per model and 5,120 primary scores in total. Both reviewers independently scored these 200 outputs for agreement analysis; the second scores do not enter the model means. Confidence intervals use 10,000 stratified paired-image percentile bootstrap resamples, with the primary reviewer assignment defining the strata.

\begin{table*}[t]
\centering
\caption{Operational anchors for the four blinded human-evaluation criteria.}
\label{tab:caption-human-rubric}
\begin{scriptsize}
\setlength{\tabcolsep}{3.5pt}
\begin{tabular}{@{}>{\raggedright\arraybackslash}p{0.14\textwidth} >{\raggedright\arraybackslash}p{0.26\textwidth} >{\raggedright\arraybackslash}p{0.26\textwidth} >{\raggedright\arraybackslash}p{0.26\textwidth}@{}}
\toprule
Criterion & Score 0 & Score 1 & Score 2 \\
\midrule
M1: Subject correctness & The main subject is wrong, replaced by an unrelated object, or omitted. & The subject is related but overly broad or only partly identified, or the fine-grained name is incompatible with the fixed class. & The main subject is correct, and any taxonomic name is compatible with the fixed class. \\
M2: Visual factuality & A major count or logical error, a fabricated object, or multiple unsupported statements change the main meaning. & The core description is correct but contains one minor or local visual error. & Claims about color, form, texture, count, arrangement, state, behavior, and scene are supported by the image. \\
M3: Grounded informativeness & No usable image information is provided, or the content is mostly incorrect. & The subject is correct, but the description is clearly generic or omits salient information. & The subject is correct and at least one discriminative, image-supported attribute or scene detail is given; a complete, accurate short sentence also qualifies for a simple image. \\
M4: Language and format & The output is empty or unusable, contains multi-sentence explanation or prompt leakage, or seriously violates the length or format requirements. & The output is understandable but has a minor grammar, punctuation, completeness, or format issue. & The output is a clear, grammatically complete English sentence of 1--30 words, with no list, explanation, line break, or prompt residue. \\
\bottomrule
\end{tabular}
\end{scriptsize}
\end{table*}

\begin{table}[htbp]
\centering
\caption{Blinded human evaluation on the 512-image captioning subset. Overall scores are primary-score means with 95\% bootstrap confidence intervals. M1--M4 are mean criterion scores on a 0--2 scale; higher values are better, and column-best means are bold.}
\label{tab:caption-human-scores}
\begin{small}
\setlength{\tabcolsep}{4.2pt}
\begin{tabular}{@{}>{\raggedright\arraybackslash}p{0.27\textwidth} c c c c c@{}}
\toprule
Model & Overall / 8 (95\% CI) & M1 & M2 & M3 & M4 \\
\midrule
Qwen3-VL-4B & \textbf{6.89 (6.78--6.99)} & 1.488 & \textbf{1.566} & \textbf{1.840} & 1.992 \\
Qwen2.5-VL-3B & 6.82 (6.69--6.96) & \textbf{1.549} & 1.535 & 1.744 & 1.996 \\
LLaVA-1.5-7B & 6.17 (6.00--6.34) & 1.354 & 1.479 & 1.447 & 1.895 \\
InternVL2.5-8B & 5.88 (5.74--6.02) & 1.109 & 1.188 & 1.604 & 1.980 \\
BLIP-Large & 5.61 (5.42--5.79) & 1.203 & 1.254 & 1.340 & 1.812 \\
Florence-2 Base FT & 5.60 (5.40--5.79) & 1.076 & 1.301 & 1.291 & 1.930 \\
BLIP-Base & 4.84 (4.62--5.05) & 0.943 & 1.084 & 1.074 & 1.736 \\
LLaVA-Med 1.5 7B & 3.25 (3.13--3.36) & 0.176 & 0.562 & 0.510 & \textbf{2.000} \\
ViT-GPT2 & 2.57 (2.46--2.69) & 0.088 & 0.252 & 0.385 & 1.848 \\
Phi-3.5 Vision & 2.51 (2.41--2.61) & 0.029 & 0.230 & 0.357 & 1.895 \\
\bottomrule
\end{tabular}
\par\smallskip
\parbox{\linewidth}{\raggedright\footnotesize\textit{Model references.}
ViT-GPT2~\citep{nlpconnect2022vitgpt2,dosovitskiy2021image,radford2019gpt2};
BLIP-Large and BLIP-Base~\citep{li2022blip}; Florence-2 Base FT~\citep{xiao2024florence2};
LLaVA-Med 1.5 7B~\citep{li2023llavamed}; Qwen2.5-VL-3B~\citep{bai2025qwen25vl};
Qwen3-VL-4B~\citep{bai2025qwen3vl}; LLaVA-1.5-7B~\citep{liu2024llava15};
InternVL2.5-8B~\citep{chen2024internvl25}; Phi-3.5 Vision~\citep{abdin2024phi3}.}
\end{small}
\end{table}

The agreement set indicates that the two reviewers most often assign exactly the same language-and-format score (92\%) and subject-correctness score (77\%). Exact agreement is 66\% for visual factuality and 63\% for grounded informativeness. Table~\ref{tab:caption-human-agreement} also reports quadratic-weighted $\kappa$, which accounts for the distance between ratings on the three-point scale. Grounded informativeness has the lowest exact agreement and a weighted $\kappa$ of 0.538, indicating greater reviewer dependence for this criterion. The 200-caption overlap set measures agreement, while model means retain one primary rating per output.

\begin{table}[htbp]
\centering
\caption{Inter-reviewer agreement on the 200 captions scored by both reviewers. These overlap scores are used only for agreement analysis.}
\label{tab:caption-human-agreement}
\begin{small}
\begin{tabular}{@{}>{\raggedright\arraybackslash}p{0.14\textwidth} >{\raggedright\arraybackslash}p{0.42\textwidth} r r@{}}
\toprule
Dimension & Criterion & Exact agreement & Weighted $\kappa$ \\
\midrule
M1 & Subject correctness & 77\% & 0.838 \\
M2 & Visual factuality & 66\% & 0.729 \\
M3 & Grounded informativeness & 63\% & 0.538 \\
M4 & Language and format & 92\% & 0.538 \\
\bottomrule
\end{tabular}
\end{small}
\end{table}

For M4, both reviewers' ratings are strongly concentrated at score 2, so the 92\% exact agreement coexists with a lower chance-corrected weighted $\kappa$.

\subsection{Complete Automatic Metric Matrix}
\label{app:caption-metrics}

\begin{figure}[htbp]
  \centering
  \hcenteredincludegraphics[width=\linewidth]{-11.2pt}{11.2pt}{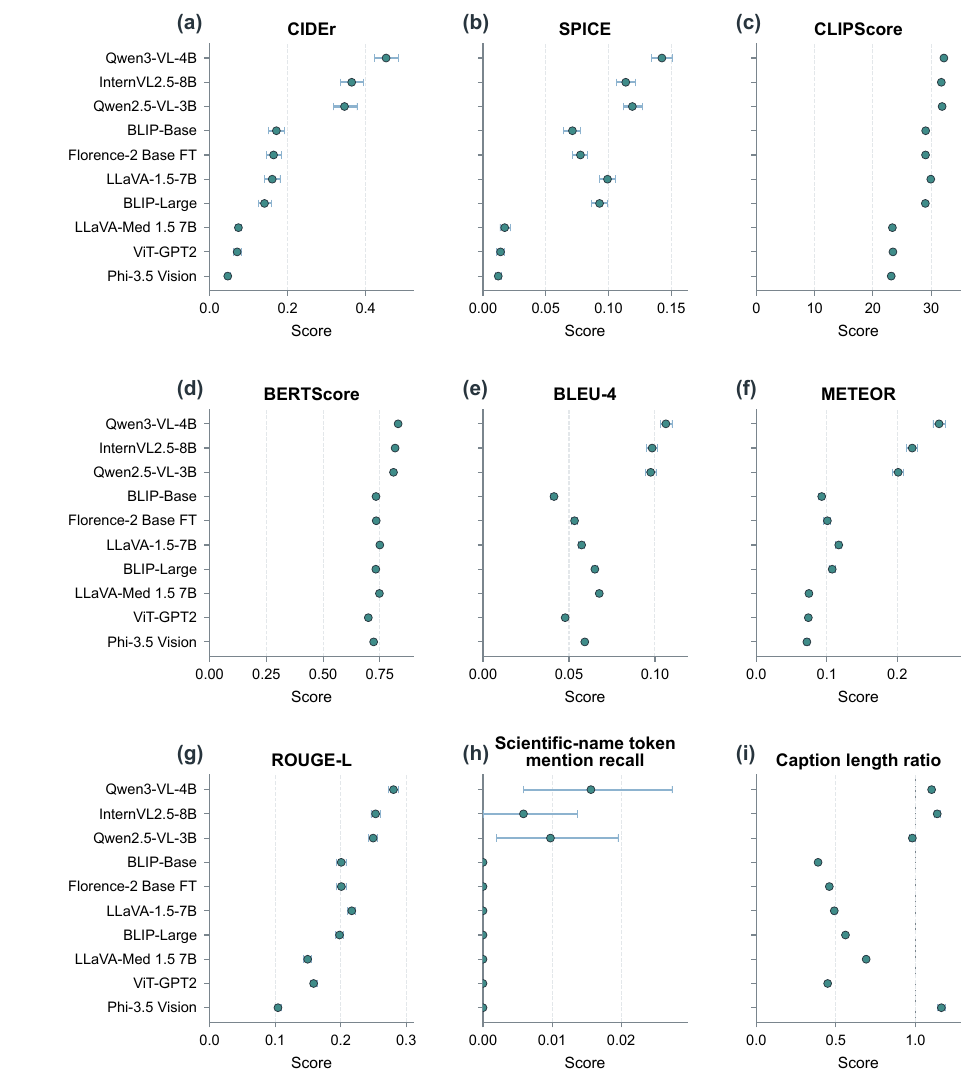}
  \caption{Complete automatic caption metrics for ten open models on 512 images. Panels report CIDEr, SPICE, CLIPScore, BERTScore, BLEU-4, METEOR, ROUGE-L, scientific-name token mention recall, and caption length ratio. Points show model estimates, and horizontal bars show image-level 95\% bootstrap intervals; the dashed line marks a length ratio of 1.0.}
  \label{fig:caption-metrics-complete}
\end{figure}

All nine automatic measures are computed for every model. CIDEr~\citep{vedantam2015cider}, SPICE~\citep{anderson2016spice}, BLEU-4~\citep{papineni2002bleu}, METEOR~\citep{banerjee2005meteor}, and ROUGE-L~\citep{lin2004rouge} compare generated text with the expert-annotated reference descriptions. BERTScore~\citep{zhang2020bertscore} measures contextual textual similarity. CLIPScore~\citep{hessel2021clipscore} measures image--text alignment. Scientific-name token mention recall records whether the final two normalized tokens of the class scientific name both occur in the generated caption. Caption length ratio compares generated and reference lengths. Figure~\ref{fig:caption-metrics-complete} reports all nine measures for all ten models, with an image-level 95\% bootstrap interval for every estimate.

CIDEr uses a single-reference TF-IDF n-gram cosine score. SPICE uses the \texttt{Spice} class from \texttt{pycocoevalcap.spice}. BERTScore reports English F1 from \texttt{distilbert-\allowbreak base-\allowbreak uncased} without baseline rescaling. CLIPScore uses \texttt{openai/\allowbreak clip-vit-\allowbreak base-\allowbreak patch32} and reports the cosine similarity between L2-normalized image and text embeddings, clipped at zero and multiplied by 100.

\subsection{Cross-Metric Observations}
\label{app:caption-observations}

Table~\ref{tab:caption-human-scores} separates content judgments from output form. In particular, the three models with the lowest overall human scores retain mean language-and-format scores of 1.85--2.00 but have subject-correctness means of 0.03--0.18. Figure~\ref{fig:caption-metrics-complete} preserves all nine automatic measures as separate scores. Reference-overlap metrics emphasize lexical and propositional agreement with the expert annotations. CLIPScore emphasizes image--text compatibility, while scientific-name token mention recall measures spontaneous full-binomial mention under a prompt that also permits common or genus names; caption length ratio describes output form. These complete-pipeline comparisons combine visual content, output length and format, and interface-specific instruction following. Ranking differences across the human and automatic measures reflect the different properties that they assess.

\clearpage
\subsection{Qualitative Failure Cases}
\label{app:caption-failure-cases}

Aggregate metrics summarize performance over the fixed subset but do not show the content of individual errors. Figures~\ref{fig:caption-failure-argopecten}--\ref{fig:caption-failure-melongena} compare five expert-annotated reference captions with the exact outputs produced by five evaluated models for images where the disagreement is visually clear. The examples include substitution of unrelated objects, confusion between a shell and another object category, omission of the main subject, and unsupported visual uncertainty.

\input{fig/appendix/figure_d2_caption_failure_cases}

\section{Extended Synthetic Sensitivity Analysis}
\label{app:synthetic-analysis}

\subsection{Complete Transformation-by-Level Results}
\label{app:synthetic-sensitivity}

\begin{figure}[htbp]
  \centering
  \hcenteredincludegraphics[width=\linewidth]{-6.3pt}{6.3pt}{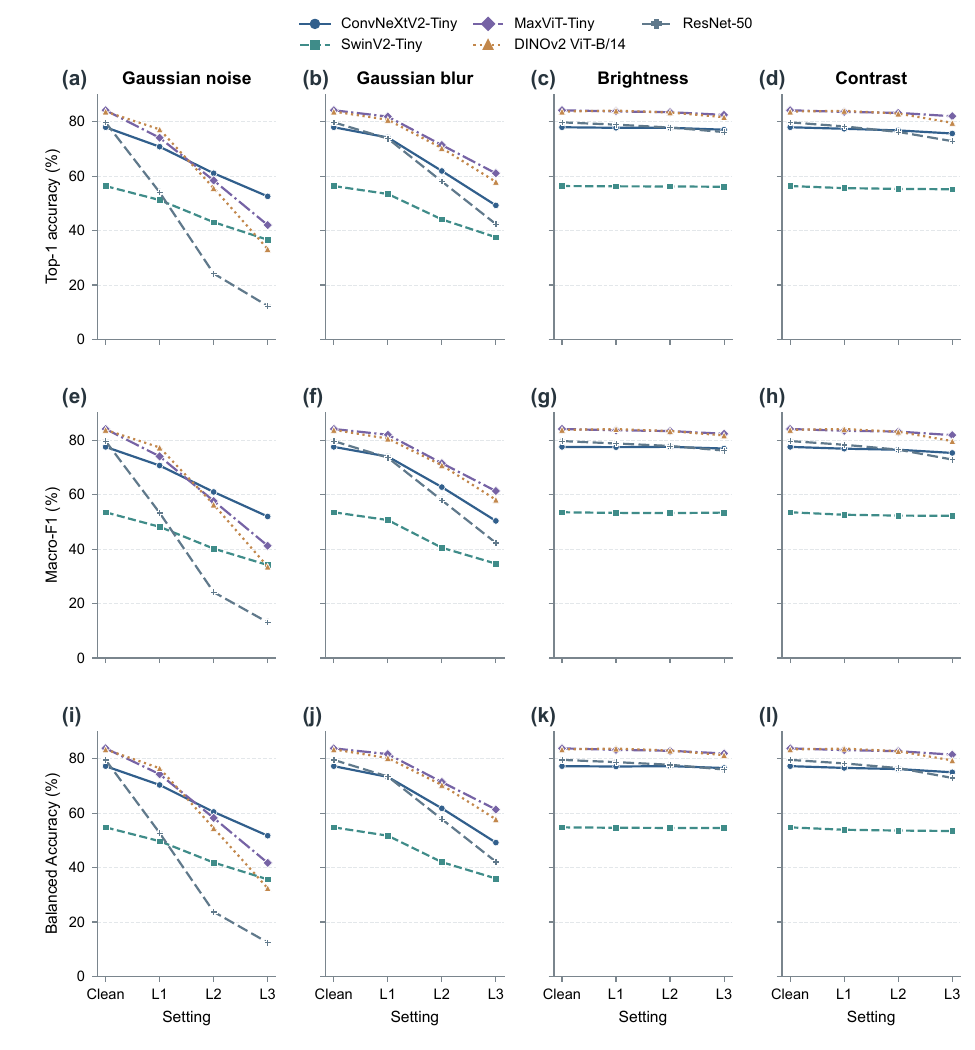}
  \caption{Complete controlled synthetic sensitivity results for five fixed checkpoints on 3,958 held-out images. Rows report Top-1, Macro-F1, and Balanced Accuracy; columns report Gaussian noise, Gaussian blur, brightness, and contrast. Each panel includes clean inputs and the three prescribed levels, with consistent checkpoint encodings across panels.}
  \label{fig:sensitivity-complete}
\end{figure}

The generic sensitivity analysis independently applies Gaussian noise, Gaussian blur, brightness changes, and contrast changes to the same 3,958 held-out images. All evaluations use fixed seed-41 checkpoints. Each transformation is applied independently to the cached $224\times224$ RGB image before model normalization. Gaussian noise uses a deterministic image-specific draw in the $[0,1]$ pixel range and is clipped to that range. Gaussian blur uses Pillow's \texttt{GaussianBlur} operator with the reported radius. Brightness and contrast use Pillow's \texttt{ImageEnhance} operators. Gaussian noise uses standard deviations 0.04, 0.08, and 0.12. Gaussian blur uses radii 0.8, 1.4, and 2.0. Brightness factors are 0.80, 0.65, and 0.50; contrast factors are 0.75, 0.55, and 0.35. Main-text Figure~\ref{Fig:synthetic-sensitivity}(a) summarizes the five-checkpoint mean Top-1 trends. Figure~\ref{fig:sensitivity-complete} reports Top-1, Macro-F1, and Balanced Accuracy for every checkpoint, transformation, and level.

The generic and marine-inspired panels use different targeted five-checkpoint sets. We interpret each average within its own panel and use the analyses to characterize the evaluated fixed checkpoints.

Across the five selected fixed checkpoints, the strongest noise and blur settings produce larger average changes than the strongest brightness and contrast settings. At level 3, mean Top-1 is 35.37\% for Gaussian noise and 49.64\% for Gaussian blur. The corresponding values are 74.71\% for brightness and 73.06\% for contrast. Macro-F1 and Balanced Accuracy follow the same transformation ordering at the strongest settings, as shown in Figure~\ref{fig:sensitivity-complete}. The ordering of the evaluated checkpoints changes with transformation strength, especially under Gaussian noise. Milder brightness and contrast changes preserve more of the clean-image ordering. The comparison distinguishes fixed checkpoints with similar held-out accuracy but different sensitivity profiles under the specified synthetic changes.

\section{Additional Frozen-Test Diagnostics}
\label{app:additional-diagnostics}

\subsection{Marine-Inspired RGB Transformations}
\label{app:marine-degradation}

This analysis applies three fixed RGB transformations to the same 3,958 held-out images. Five fixed seed-41 checkpoints are evaluated without retraining or test-time augmentation. The two water-attenuation conditions use Jerlov II and Jerlov III coefficients from the released CADDY alpha-blend implementation~\citep{gomezchavez2019caddy,gomezchavez2019underwaterdistortions}. For channel $c$, the attenuation mask is $255\exp(-K_{d,c}\cdot5)$. The transformation uses an equal blend of the input and mask. This empirical additive RGB color-cast and contrast-compression stress test is adapted from the CADDY utility; it is not a depth-resolved radiative-transfer simulation. The Jerlov II BGR coefficients are $(0.0619,0.0998,0.580)$; Jerlov III uses $(0.117,0.145,0.616)$. Under this formula, the larger Jerlov II mask transmission terms yield a brighter additive mask; the observed Jerlov II--III ordering is therefore not interpreted as a physical severity ranking of water types. The marine-snow condition adapts the PHISWID marine-snow synthesis to the RGB-only benchmark inputs~\citep{kaneko2026phiswid}.

The five-checkpoint mean Top-1 is 72.66\% on clean images. It is 49.07\% under Jerlov II attenuation, 55.84\% under marine snow, and 51.20\% under Jerlov III attenuation. Main-text Figure~\ref{Fig:synthetic-sensitivity}(b) reports the corresponding clean-to-transformed decreases for each checkpoint, while Figure~\ref{fig:marine-long-tail}(a) retains the absolute Top-1 results.

\subsection{Class-Frequency Imbalance}
\label{app:class-frequency-imbalance}

Training support ranges from 106 to 401 images per class. The maximum-to-minimum ratio is 3.783, and the training-support Gini coefficient is 0.1395. ShellfishNet therefore has moderate class-frequency imbalance rather than an extreme long-tail distribution. We define support-ranked strata before inspecting test metrics. Tail contains the lowest 33 classes, Medium the next 33, and Head the highest 34. These groups account for 22.00\%, 36.37\%, and 41.62\% of the training images.

\begin{figure}[htbp]
  \centering
  \hcenteredincludegraphics[width=\linewidth]{-14.8pt}{14.8pt}{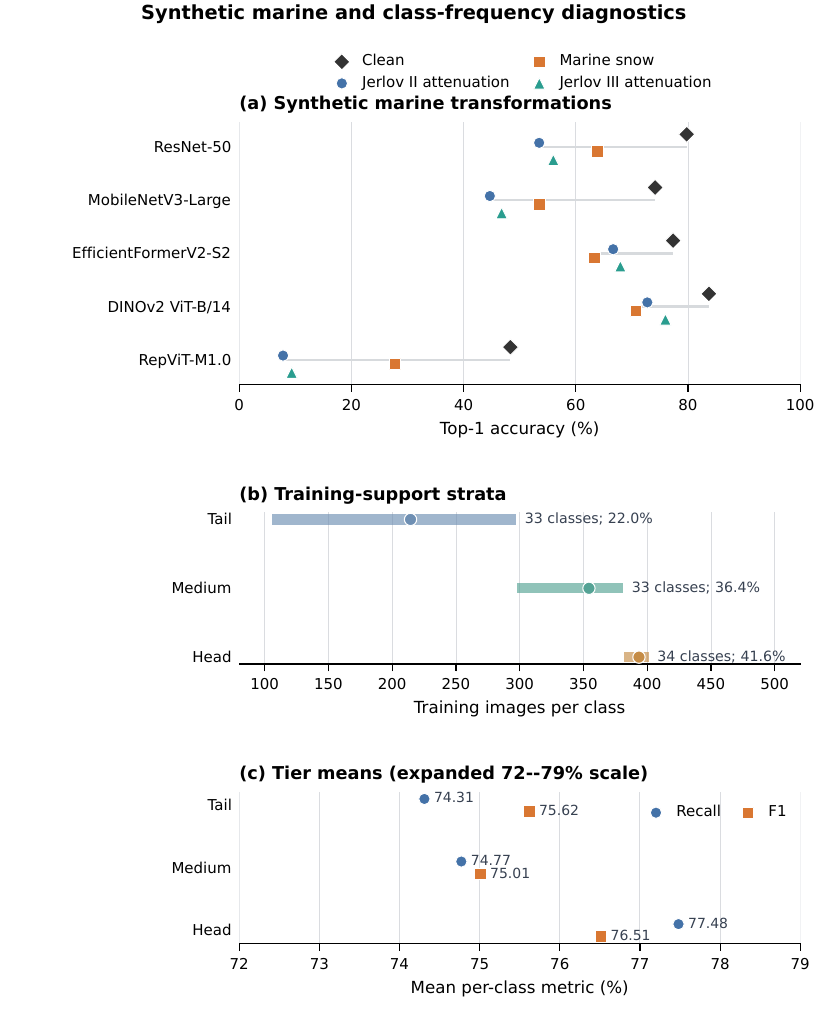}
  \caption{Marine-inspired RGB transformations and class-frequency analysis. Panel (a) reports absolute Top-1 accuracy for clean images and three fixed transformations, using five fixed seed-41 checkpoints on the same 3,958-image test partition. Panel (b) shows the training-support range and mean for each support-ranked stratum; labels give class counts and training-image shares. Panel (c) reports mean per-class Recall and F1 across all 69 model--seed runs on an expanded 72--79\% scale.}
  \label{fig:marine-long-tail}
\end{figure}

The performance summary covers all 69 formal model--seed runs and 6,900 per-class records. Mean per-class Recall is 74.31\% for Tail, 74.77\% for Medium, and 77.48\% for Head. Mean F1 is 75.62\%, 75.01\%, and 76.51\%, respectively. Figure~\ref{fig:marine-long-tail} reports the support strata and both metrics. No resampling, reweighting, augmentation, or retraining is used in this analysis. The stratum differences describe an association with training support; they are not a causal estimate of sample-count effects.

%% file: fig/appendix/figure_d2_caption_failure_cases.tex
\begin{figure}[H]
\centering
\begin{minipage}[c]{0.32\linewidth}
\centering
\hcenteredincludegraphics[width=\linewidth]{0pt}{0pt}{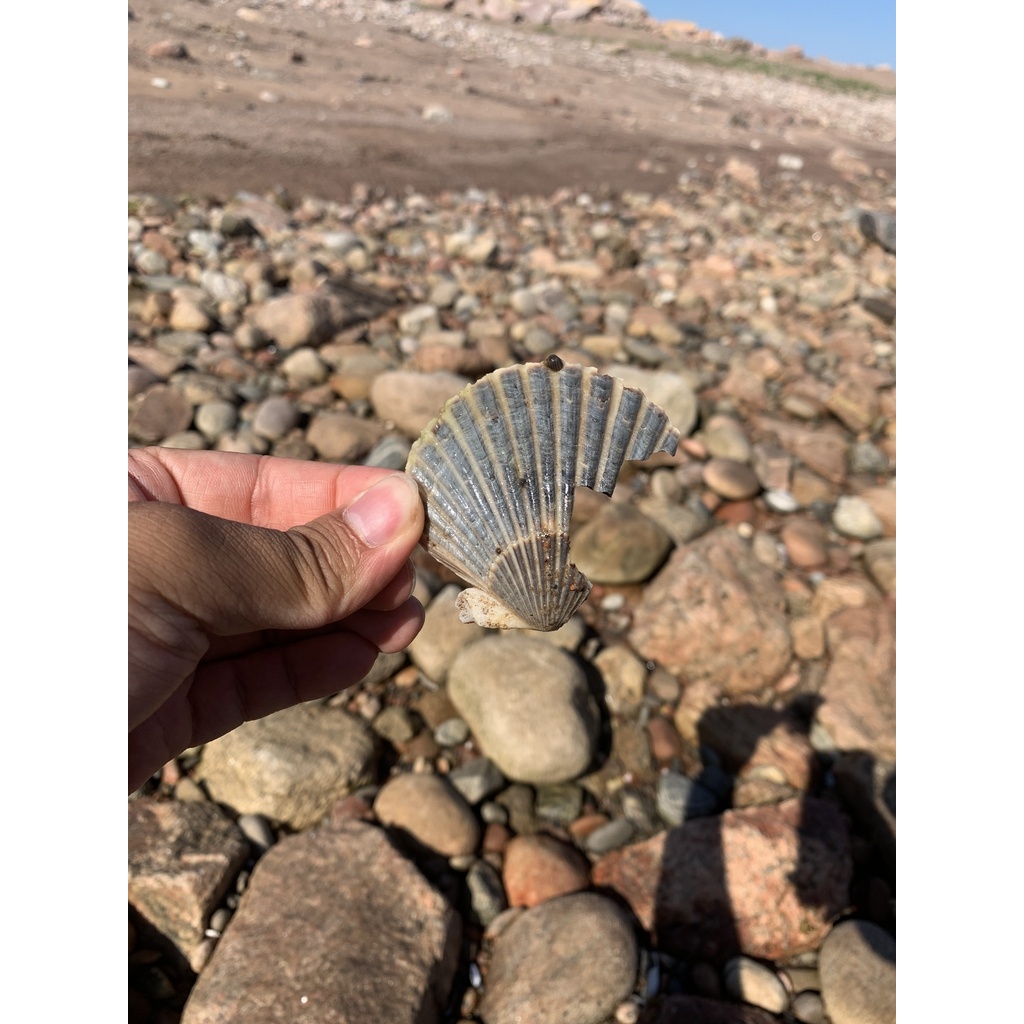}
\end{minipage}\hfill
\begin{minipage}[c]{0.65\linewidth}
\small
\textbf{Ground Truth (Expert Annotation):} \\
A bay scallop (\textit{Argopecten irradians}) shell, gray-brown and fan-shaped with deep radial ribs, is held above a rocky intertidal shore.

\vspace{0.8em}
\hrule
\vspace{0.8em}

\textbf{ViT-GPT2 (Failure Case):} \\
a person holding a small bird on a beach
\end{minipage}
\caption{Qualitative comparison for \textit{Argopecten irradians}. ViT-GPT2 replaces the clearly visible hand-held fan-shaped shell with an unrelated bird.}
\label{fig:caption-failure-argopecten}
\end{figure}

\begin{figure}[H]
\centering
\begin{minipage}[c]{0.32\linewidth}
\centering
\hcenteredincludegraphics[width=\linewidth]{0pt}{0pt}{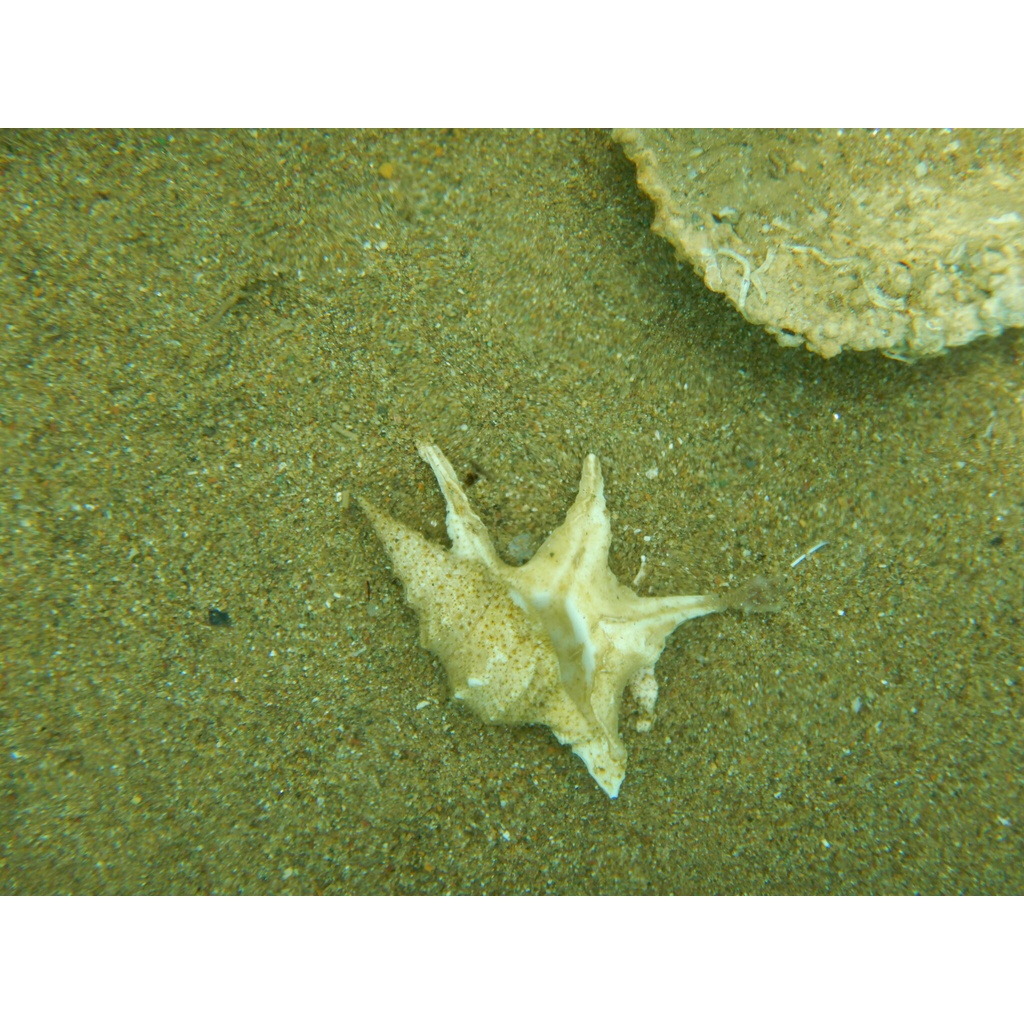}
\end{minipage}\hfill
\begin{minipage}[c]{0.65\linewidth}
\small
\textbf{Ground Truth (Expert Annotation):} \\
A pale, spiny \textit{Aporrhais pespelecani} shell rests on fine sandy sediment in a shallow underwater marine scene.

\vspace{0.8em}
\hrule
\vspace{0.8em}

\textbf{BLIP-Large (Failure Case):} \\
there is a starfish that is laying on the sand
\end{minipage}
\caption{Qualitative comparison for \textit{Aporrhais pespelecani}. BLIP-Large mistakes the prominent spiny shell for a starfish.}
\label{fig:caption-failure-aporrhais}
\end{figure}

\begin{figure}[H]
\centering
\begin{minipage}[c]{0.32\linewidth}
\centering
\hcenteredincludegraphics[width=\linewidth]{0pt}{0pt}{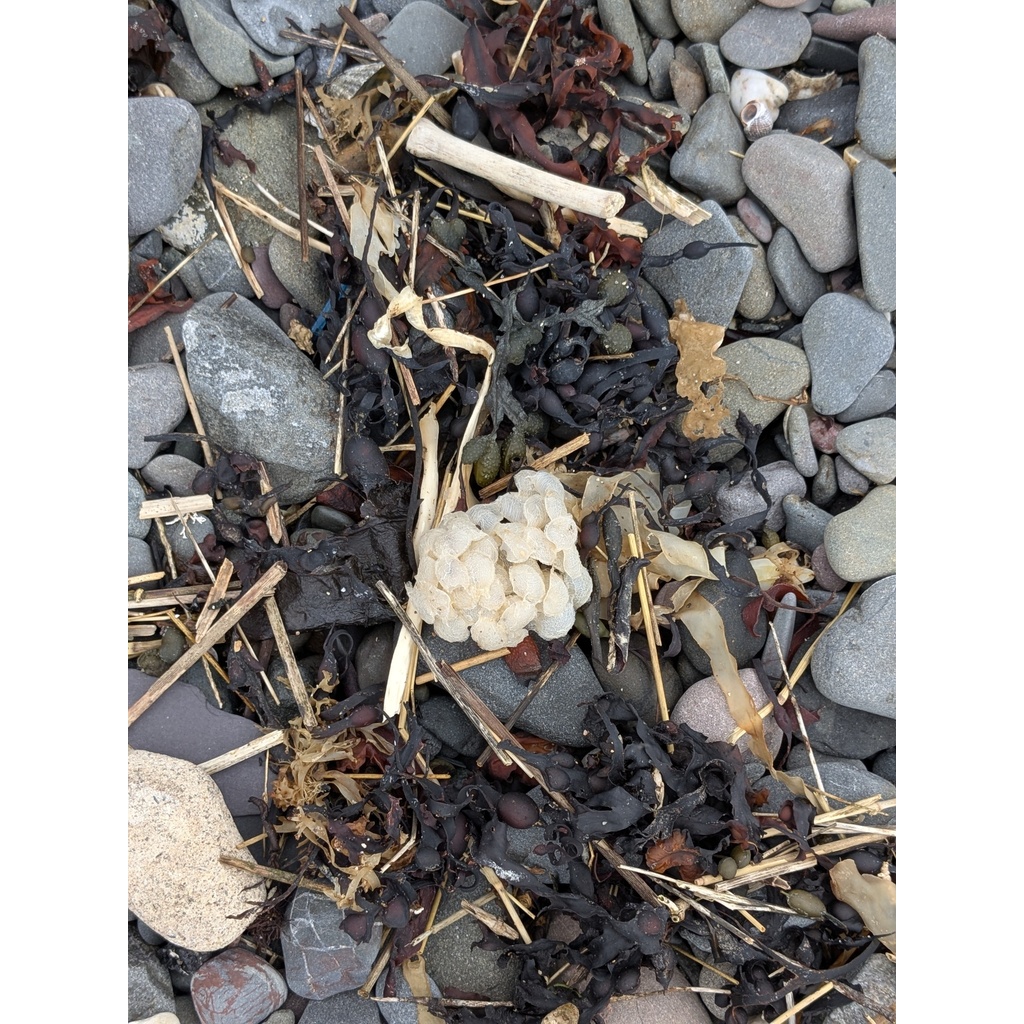}
\end{minipage}\hfill
\begin{minipage}[c]{0.65\linewidth}
\small
\textbf{Ground Truth (Expert Annotation):} \\
A pale cluster of common whelk (\textit{Buccinum undatum}) egg capsules rests amid seaweed and driftwood on rounded beach stones at the tide line.

\vspace{0.8em}
\hrule
\vspace{0.8em}

\textbf{Florence-2 Base FT (Failure Case):} \\
A pile of rocks and twigs on the ground.
\end{minipage}
\caption{Qualitative comparison for \textit{Buccinum undatum} egg capsules. Florence-2 Base FT describes the surrounding stones and twigs but omits the central pale cluster.}
\label{fig:caption-failure-buccinum}
\end{figure}

\begin{figure}[H]
\centering
\begin{minipage}[c]{0.32\linewidth}
\centering
\hcenteredincludegraphics[width=\linewidth]{0pt}{0pt}{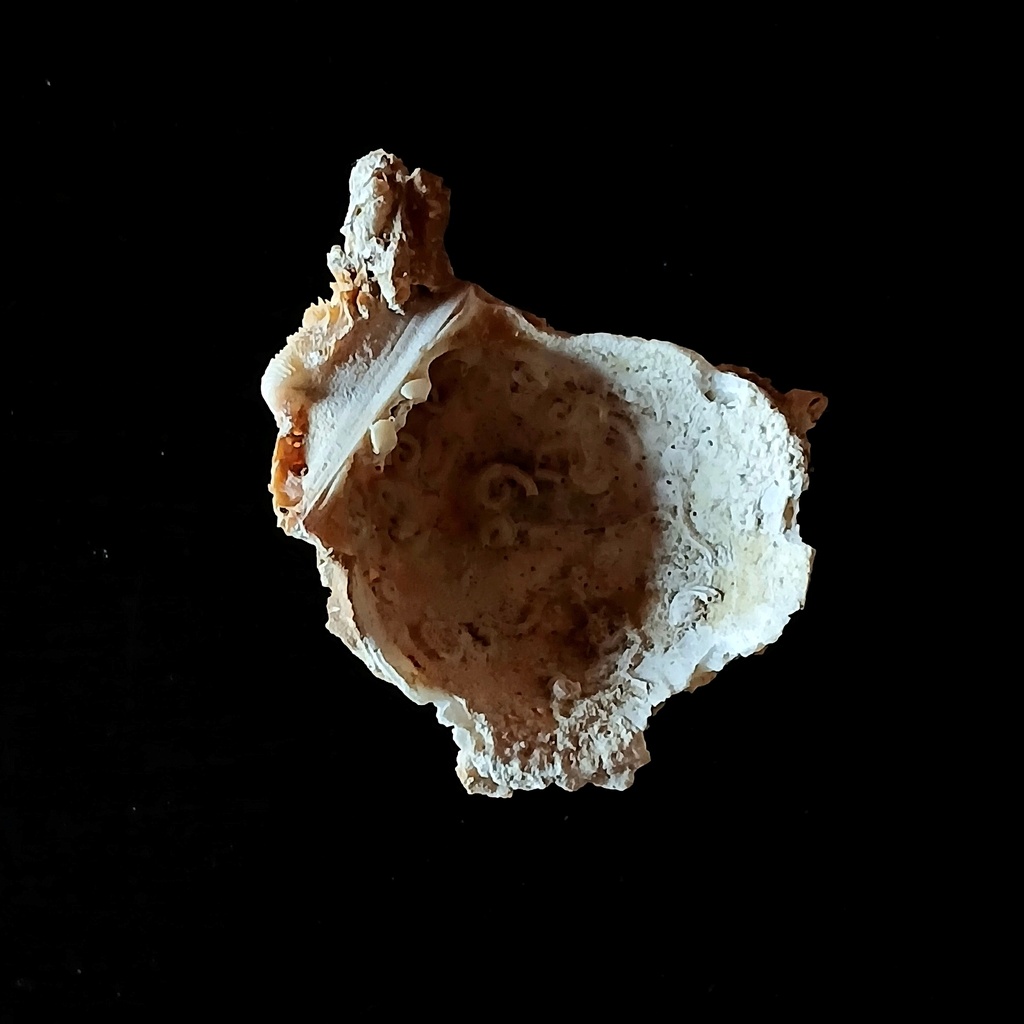}
\end{minipage}\hfill
\begin{minipage}[c]{0.65\linewidth}
\small
\textbf{Ground Truth (Expert Annotation):} \\
A weathered, irregular \textit{Spondylus gaederopus} shell, with jagged projections and chalky white-to-brown texture, is shown as an isolated marine bivalve specimen against a dark background.

\vspace{0.8em}
\hrule
\vspace{0.8em}

\textbf{LLaVA-1.5-7B (Failure Case):} \\
A rock with a hole in it sits on a black surface.
\end{minipage}
\caption{Qualitative comparison for \textit{Spondylus gaederopus}. LLaVA-1.5-7B mistakes the isolated jagged shell for a rock.}
\label{fig:caption-failure-spondylus}
\end{figure}

\begin{figure}[H]
\centering
\begin{minipage}[c]{0.32\linewidth}
\centering
\hcenteredincludegraphics[width=\linewidth]{0pt}{0pt}{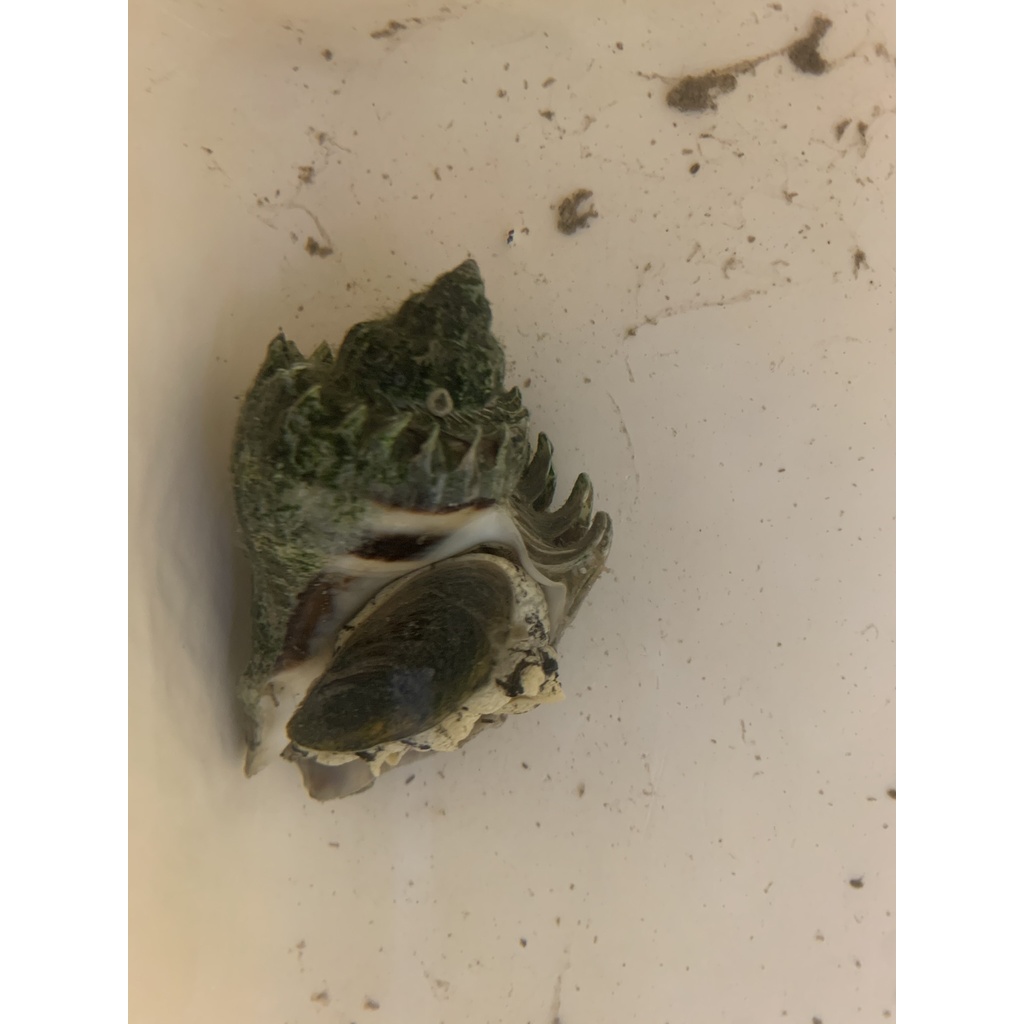}
\end{minipage}\hfill
\begin{minipage}[c]{0.65\linewidth}
\small
\textbf{Ground Truth (Expert Annotation):} \\
A spiny, algae-darkened crown conch (\textit{Melongena corona}) rests in shallow seawater on a pale sandy surface scattered with fine marine debris.

\vspace{0.8em}
\hrule
\vspace{0.8em}

\textbf{Phi-3.5 Vision (Failure Case):} \\
The image shows a blurred and indistinct marine scene, possibly a painting, with no clear subject or details to identify.
\end{minipage}
\caption{Qualitative comparison for \textit{Melongena corona}. Phi-3.5 Vision reports a blurred, indeterminate scene despite the clearly visible spiny shell.}
\label{fig:caption-failure-melongena}
\end{figure}